\documentclass[journal]{IEEEtran}

\usepackage{amsmath,amssymb}
\usepackage{graphicx}
\usepackage{booktabs}
\usepackage{tabularx,array}
\usepackage{url}
\usepackage{xcolor}
\usepackage[hidelinks]{hyperref}
\definecolor{demolink}{HTML}{007C83}
\title{AGE-Capture: Autonomous, General, and Efficient UAV Capture for High-Fidelity Scene Reconstruction}

\newif\ifageshowauthors
\ageshowauthorstrue
\ifageshowauthors
  \author{%
    Xiaofeng Jin\textsuperscript{1,\ensuremath{\dagger}},
    Ningbo Bu\textsuperscript{2,\ensuremath{\dagger}},
    Matteo Frosi\textsuperscript{1},
    Yiran Guo\textsuperscript{1},
    Weilun Ren\textsuperscript{3},
    and Matteo Matteucci\textsuperscript{1}\\[0.4em]
    {\small
      \textsuperscript{1}~Politecnico di Milano \quad
      \textsuperscript{2}~University of Chinese Academy of Sciences \quad
      \textsuperscript{3}~Zhejiang University}%
    \thanks{\textsuperscript{\ensuremath{\dagger}}~%
  Xiaofeng Jin and Ningbo Bu contributed equally to this work.
  E-mail (Xiaofeng Jin):
  \href{mailto:xiaofeng.jin@polimi.it}{\nolinkurl{xiaofeng.jin@polimi.it}}.}%
  }
\else
  \author{}
\fi

\begin{document}
\IEEEaftertitletext{%
  \begin{center}
    \includegraphics[width=\textwidth]{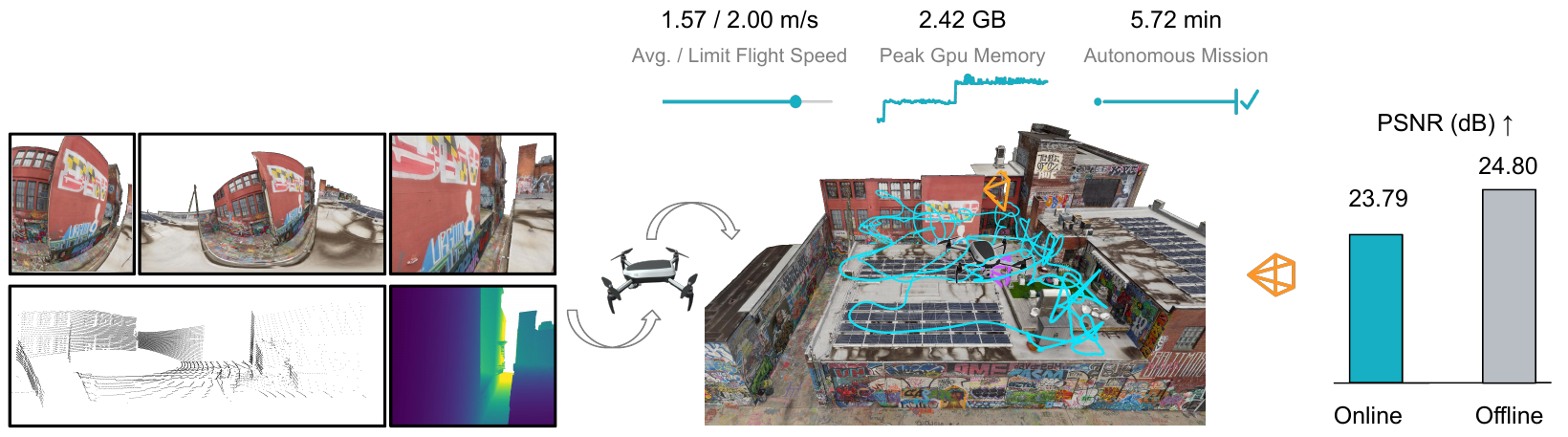}\par\vspace{1mm}%
    \parbox{\textwidth}{\footnotesize\refstepcounter{figure}\label{fig:teaser}Fig.~\thefigure. AGE-Capture for high-fidelity UAV scene reconstruction. Complementary RGB and range sensing supports autonomous reconstruction-oriented acquisition; the resulting observations provide an online reconstruction and a compact view set for offline reconstruction. The example reports average flight speed relative to the configured limit, peak GPU memory, autonomous mission duration, and online/offline peak signal-to-noise ratio (PSNR).\enspace
\href{https://polimi365-my.sharepoint.com/:v:/g/personal/11092598_polimi_it/IQCOsuFmdrp5SqZyeuQfXVkvAXOEQASo0H_guSznU5dS8Fo?nav=eyJyZWZlcnJhbEluZm8iOnsicmVmZXJyYWxBcHAiOiJPbmVEcml2ZUZvckJ1c2luZXNzIiwicmVmZXJyYWxBcHBQbGF0Zm9ybSI6IldlYiIsInJlZmVycmFsTW9kZSI6InZpZXciLCJyZWZlcnJhbFZpZXciOiJNeUZpbGVzTGlua0NvcHkifX0&e=6gcenh}{\textcolor{demolink}{\textbf{[Video]}}}}\normalsize
  \end{center}%
}
\maketitle

\begin{abstract}
High-fidelity scene reconstruction with an unmanned aerial vehicle (UAV) requires more than identifying valuable viewpoints: the system must autonomously acquire informative observations and determine task completion under different sensing configurations and limited flight resources.
We present AGE-Capture, an autonomous, sensor-adaptive, and efficient active-capture framework for real UAVs.
Building on renderability, the framework uses a lightweight online state to evaluate the utility of candidate views, assess task progress, and guide flight execution.
By decoupling geometric exploration from reconstruction observations and establishing a unified camera-ray interface, the framework allows different range sensors and RGB camera models to share a reconstruction-driven acquisition pipeline.
Reconstruction-frontier viewpoints characterize residual acquisition value for region screening and autonomous termination; dynamics-aware hierarchical planning combined with asynchronous event-driven capture reduces unnecessary inter-region motion, time spent waiting for plans, and repeated observations.
Across ten indoor and outdoor simulation scenes, AGE-Capture achieves a mean peak signal-to-noise ratio (PSNR) $3.30$--$3.53\,\mathrm{dB}$ higher than that of the online baselines while using $58.2\%$ of the common mission-time budget on average. Its mean peak GPU memory usage is $91.0\%$--$93.3\%$ lower than that of the same baselines.
Experiments with different sensor configurations validate the adaptability of the shared acquisition pipeline, while flights in nine real-world scenes further demonstrate onboard closed-loop operation under practical computational and motion constraints.
\end{abstract}

\section{Introduction}
\label{sec:introduction}

Neural rendering and Gaussian Splatting have advanced high-fidelity 3D reconstruction and novel-view synthesis~\cite{mildenhall2020nerf,kerbl20233dgaussians}. Yet reconstruction quality remains strongly dependent on the spatial and angular coverage of the acquired observations. Active reconstruction addresses this dependence by selecting observation positions and viewing directions that improve the current scene representation under limited acquisition resources. For large-scale real-world scenes, UAVs provide a practical platform because they can reach viewpoints at different elevations and headings (Fig.~\ref{fig:teaser}).

Existing active-capture methods primarily address where to observe next. Traditional next-best-view (NBV) methods evaluate candidate viewpoints using geometric uncertainty, information gain, coverage, or visibility~\cite{bircher2016receding}. More recent approaches estimate the potential value of candidate observations through neural-field uncertainty, radiance-field feedback, or related reconstruction-aware signals, thereby relating observation utility more directly to the reconstruction objective~\cite{lee2022uncertainty,jin2023neunbv,nvf2024}.

Identifying valuable candidate viewpoints, however, is only one part of a deployable UAV active-reconstruction system. The system must also accommodate heterogeneous sensing configurations and coordinate acquisition decisions with execution under limited flight resources. Existing methods commonly evaluate reconstruction under a fixed view count or time budget. These limits constrain capture but do not indicate whether further observations would improve the reconstruction. The system must therefore assess residual reconstruction value online and terminate capture when further acquisition is no longer warranted.

Renderability-based reconstruction gain provides a state representation suitable for this online decision process. R3CON represents the value of candidate observations through a lightweight renderability state that is decoupled from a particular reconstruction backend, avoiding repeated optimization of a complete neural radiance field during planning~\cite{r3con}. Its computation and storage requirements support frequent online queries and updates. We build a closed-loop UAV capture process around this state: heterogeneous RGB observations are converted to a common camera-ray representation and update the state; the state guides region scheduling and local NBV selection; and the UAV acquires the next observation at the selected viewpoint.

The main contributions of this work are as follows:
\begin{itemize}
    \item \textbf{A unified active reconstruction architecture for heterogeneous sensing configurations.} We decouple geometric sensing from reconstruction observations and construct a common observation interface using camera rays, allowing RGB camera models with calibrated ray mappings to share the same reconstruction-aware active capture pipeline, while the geometric exploration module handles depth-map and LiDAR inputs.
    \item \textbf{A reconstruction-frontier mechanism for autonomous task completion.} We define reconstruction-frontier viewpoints at the position level and use seed-based region screening to assess task progress and residual acquisition value, allowing the system to determine whether to continue capture under the defined reconstruction-value threshold.
    \item \textbf{An efficient active capture system under limited flight resources.} Hierarchical planning coordinates region scheduling with local view selection. Asynchronous perception, state updates, planning, and execution reduce the time spent waiting for plans.
\end{itemize}

\section{Related Work}
\label{sec:related-work}

\subsection{Viewpoint Utility for Active Reconstruction}
\label{sec:related-geometric-to-reconstruction-aware}

Early active perception and NBV planning primarily focused on geometric mapping and exploration of unknown space. Frontier-based methods prioritize the boundary between known and unknown space, whereas information-driven methods rank candidate viewpoints using criteria such as expected entropy reduction, voxel visibility, or surface coverage~\cite{yamauchi1997frontier,bourgault2002information,stachniss2005information,isler2016information,bircher2016receding}. These works established the basic active-observation paradigm: estimating the utility of candidate observations from the current scene state and selecting the view with the highest expected return.

With high-fidelity scene representations such as neural radiance fields and Gaussian Splatting entering online mapping, the focus has further shifted from geometric coverage to observation value that is more directly associated with final rendering or reconstruction quality. ActiveNeRF and NeurAR use neural-field uncertainty to assess the utility of candidate observations~\cite{activenerf2022,neurar2023}, while FisherRF uses Fisher information to characterize the potential contribution of new observations to radiance-field parameter estimation~\cite{fisherrf2024}. NVF and NARUTO further combine visibility, uncertainty, and target observation quality for active mapping and reconstruction~\cite{nvf2024,naruto2024}. Subsequent Gaussian Splatting methods construct reconstruction-aware gains from Gaussian statistics, rendering feedback, localization--exploration coupling, and multi-level information value to improve geometric and appearance reconstruction quality~\cite{agslam2024,activegs2025,gaussmi2025,activesplat2025,activegamer2025,active3d2026,activepolicy2026}.

\subsection{Motion and Mission Planning}
\label{sec:related-embodied-mission}

Beyond candidate-view valuation, active reconstruction must plan feasible motions to reach observation goals and acquire the corresponding measurements. NeurAR integrates uncertainty-driven view selection with online reconstruction, while EVPP combines implicit and volumetric representations with graph-search-based view-path planning~\cite{neurar2023,zeng2023efficient}. Other work incorporates frontier exploration into task progression for implicit reconstruction~\cite{zeng2024autonomous}. Together, these methods move reconstruction-aware planning beyond candidate-view ranking toward embodied acquisition coupled with motion, paths, and online reconstruction.

For large-scale tasks, making an individual viewpoint executable is insufficient; the system must also organize continuous observations and mission progress. TARE combines detailed local planning with sparse global planning~\cite{cao2021tare}. Informative UAV path planning combines information value, topology, and path cost~\cite{efficientipp2022}, while kinodynamic exploration incorporates vehicle constraints into target and trajectory evaluation~\cite{mendes2026kinodynamic}. TRACE directly optimizes sensor-footprint-aware trajectories in a dynamic information field~\cite{zheng2026trace}, extending discrete NBV selection to trajectory-level active reconstruction.

Their tasks are commonly bounded by a prescribed view count, path length, time budget, or exploration condition.

\subsection{Requirements for Autonomous UAV Capture}
\label{sec:related-gap}

Reconstruction-aware methods determine which observations can improve reconstruction, but their scores are often tied to a particular representation. Embodied planners organize visits to candidate viewpoints, but commonly optimize paths or trajectories under a given sensing setup and mission boundary. Autonomous UAV active reconstruction additionally requires a common reconstruction-oriented state that accepts heterogeneous RGB camera rays and supports observation-value queries, regional task progress, and completion decisions. Planning must also coordinate online state updates with trajectory execution to maintain continuous acquisition.

R3CON represents online reconstruction value with a lightweight, explicit renderability state that supports efficient queries and updates~\cite{r3con}. We use this state as feedback for closed-loop UAV capture.

\section{Method Overview}
\label{sec:method-overview}

\begin{figure*}[!t]
  \centering
  \includegraphics[width=\textwidth]{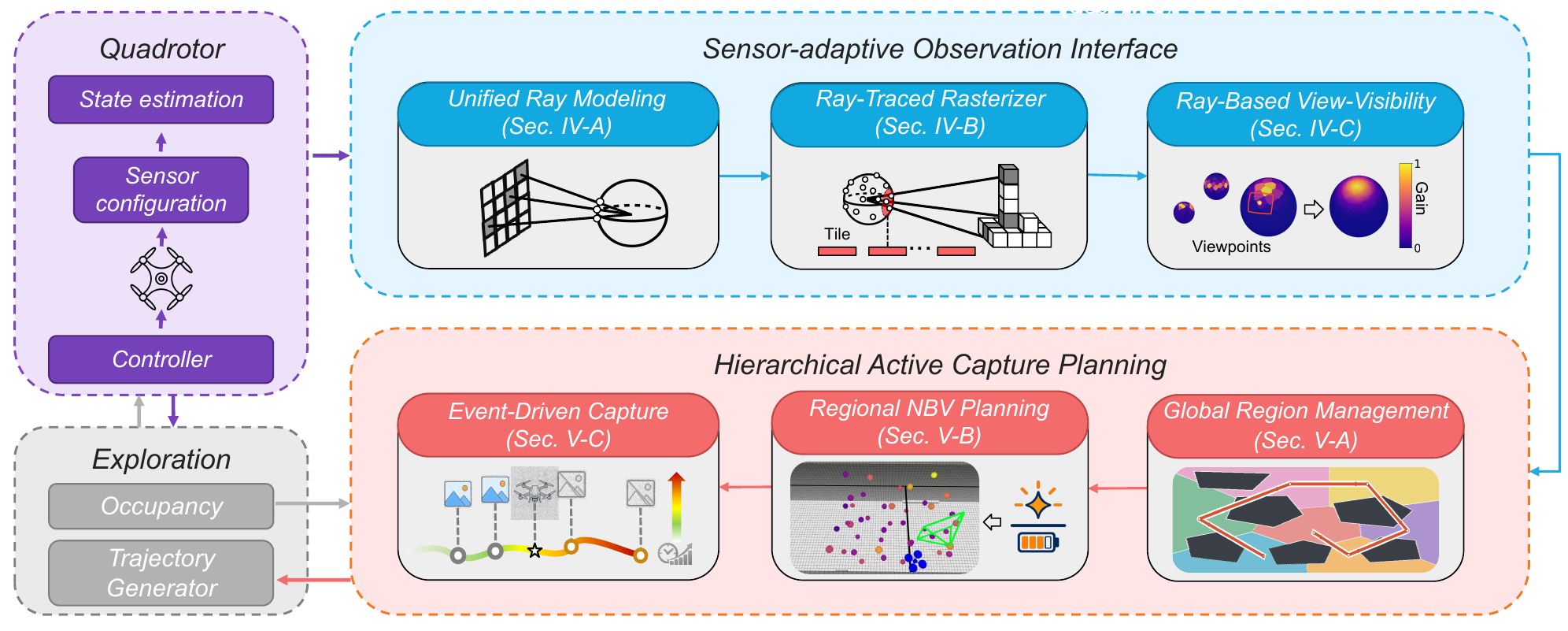}
  \caption{AGE-Capture system overview. Camera-specific RGB rays enter a shared observation interface for unified ray modeling, ray-traced renderability rasterization, and ray-based view visibility. Global region management, regional NBV planning, and event-driven capture then use the renderability state to select executable viewpoints and schedule capture, while the external exploration and trajectory modules provide geometry, motion generation, and closed-loop execution.}
  \label{fig:method-framework}
\end{figure*}

We formulate active reconstruction as a closed-loop capture process for a UAV. The system centers on a persistent renderability state that summarizes reconstruction support from acquired observations and provides residual observation value for subsequent capture. The active-reconstruction node queries this state to select an executable next-best view (NBV); executing the NBV produces the observation that updates the state and closes the loop.

Figure~\ref{fig:method-framework} shows the data flow. The exploration module maintains a geometric map, provides candidate positions and motion information, and supports region partitioning. Timestamp-matched RGB--pose observations are represented by world-space camera rays and update the persistent renderability state. During planning, the node queries this state for the gain of candidate positions and directions, and returns an NBV with its region and execution information.

The system organizes this loop into observation and planning layers. The \emph{sensor-adaptive observation interface} maps RGB observations from pinhole, fisheye, and panoramic cameras to a common world-space ray representation for renderability operations. Depth maps and LiDAR provide the geometric information used for free-space reasoning, candidate generation, and motion feasibility.

The \emph{hierarchical active capture planner} organizes decisions from scene-level task management to regional NBV selection and trajectory execution. Based on task progress, global region management selects a region for planning. Within that region, Regional NBV Planning combines renderability gain, free-space reachability, and motion cost to select a position and direction. Continuous event-driven capture registers the target observation time for the selected NBV and matches it to streaming synchronized RGB--pose data; the matched observation updates the persistent state while the next decision is prepared during execution.

\subsection{Renderability Preliminaries}
\label{sec:preliminaries}

We adopt R3CON's online renderability state to track observation support at surface locations~\cite{r3con}. For an observed surface location \(x\), its state is
\begin{equation}
    \mathcal{R}_x
    =
    \left(
    \mathcal{A}_x,\mathcal{D}_x,d_x
    \right),
    \qquad
    u(x)=\mathcal{U}(\mathcal{R}_x),
    \label{eq:renderability-state}
\end{equation}
where \(\mathcal{A}_x\) is the appearance statistic accumulated from RGB observations, \(\mathcal{D}_x\) is the set of recorded viewing-ray directions, and \(d_x\) is the minimum observation distance. The R3CON query \(\mathcal{U}\) converts these statistics into a local residual renderability score~\cite{r3con}. Timestamp-matched RGB--pose observations update this state through their camera rays and visible surfaces; its queried gains guide regional NBV selection.

\section{Sensor-Adaptive Observation Interface}
\label{sec:observation-interface}

RGB observations enter the reconstruction-oriented observation process through calibrated camera rays. Each RGB pixel and its camera pose define a world-space ray for visible-surface queries, renderability-state updates, and candidate-view evaluation. Depth maps and LiDAR provide the geometric information used for free-space reasoning, candidate generation, and motion feasibility.

\subsection{Unified Ray Modeling}
\label{sec:unified-ray-modeling}

For an RGB camera model $\mathcal{C}$, we represent every image sample $\mathbf{u}=(x,y)$ by a unit ray direction in the camera frame. Given the sampling domain $\Omega$, the camera adapter constructs the dense ray set
\begin{equation}
\mathcal{D}_{\mathcal{C}}
=
\left\{
\mathbf{d}_{c}(\mathbf{u})
\right\}_{\mathbf{u}\in\Omega},
\qquad
\mathbf{d}_{c}(\mathbf{u})
=
\frac{\Pi_{\mathcal{C}}^{-1}(\mathbf{u})}
{\left\|\Pi_{\mathcal{C}}^{-1}(\mathbf{u})\right\|_2},
\label{eq:camera_ray_workspace}
\end{equation}
where $\Pi_{\mathcal{C}}^{-1}$ denotes the inverse projection associated with $\mathcal{C}$. The inverse projection for each camera model is given by
\begin{equation}
\Pi_{\mathcal{C}}^{-1}(\mathbf{u})=
\begin{cases}
\left[
\dfrac{x-c_x}{f_x},
\dfrac{y-c_y}{f_y},
1
\right]^{\mathsf T},
& \text{pinhole},\\[10pt]
\mathbf{d}_{\mathrm{calib}}(\mathbf{u}),
& \text{wide-angle},\\[6pt]
\mathbf{d}_{\mathrm{pano}}(\mathbf{u}),
& \text{panoramic}.
\end{cases}
\label{eq:camera_model_inverse_projection}
\end{equation}
Here, $\mathbf{d}_{\mathrm{calib}}$ denotes the calibrated inverse mapping for a fisheye or other calibrated wide-angle camera, while $\mathbf{d}_{\mathrm{pano}}$ directly maps a panoramic sample to a unit-sphere direction. Normalizing the inverse-projection vector yields a unit ray direction in the camera frame.

In implementation, the camera adapter precomputes the camera-frame ray set together with the correspondence between rays and image samples. For each observation, the RGB camera pose supplies the transformation from the camera frame to the world frame, yielding the world-space ray
\begin{equation}
\mathbf{r}_{w}(\mathbf{u})
=
\left(\mathbf{o}_{w},\mathbf{R}_{wc}\mathbf{d}_{c}(\mathbf{u})\right),
\label{eq:world_ray_from_camera_ray}
\end{equation}
where $\mathbf{o}_{w}$ is the camera center and $\mathbf{R}_{wc}$ is the camera-to-world rotation. The workspace preserves the correspondence between each image sample and its ray for subsequent surface queries and renderability updates.

\subsection{Ray-Traced Renderability Rasterizer}
\label{sec:ray-traced-rasterizer}

\begin{figure}[!t]
  \centering
  \includegraphics[width=\columnwidth]{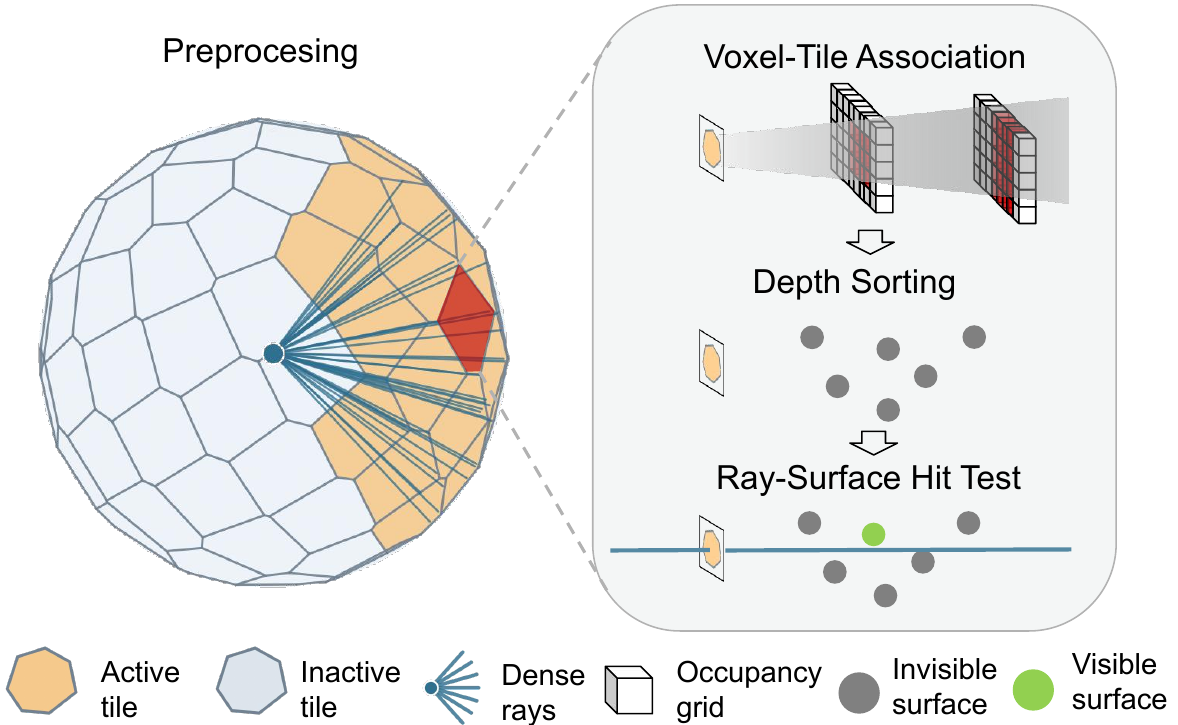}
  \caption{Ray-traced renderability rasterization. Dense observation rays are assigned to spherical Fibonacci tiles, activating only the tiles covered by the current camera. Occupancy voxels are associated with active tiles and sorted by depth before parallel ray--surface tests return the first visible surface.}
  \label{fig:ray-traced-rasterizer}
\end{figure}

Renderability evaluation requires the visible scene surface associated with each observation ray. The ray-traced rasterizer organizes visibility queries around ray directions and uses a tile-based GPU execution pattern for online processing.

The dense ray workspace is partitioned into spherical tiles whose centers are sampled on a Fibonacci lattice~\cite{gonzalez2010fibonacci}. For the $j$-th ray direction $\mathbf{d}_j$, its tile index is determined by
\begin{equation}
\kappa(j)
=
\underset{k}{\arg\max}\;
\mathbf{d}_j^{\mathsf T}\mathbf{c}_k,
\label{eq:ray_to_spherical_tile}
\end{equation}
where $\mathbf{c}_k$ is the center of spherical tile $k$. Tiles containing observation rays form the active set. Voxel primitives are assigned to tile-local candidate lists by angular overlap, sorted by depth, and processed in parallel on the GPU.

For the $j$-th ray with origin $\mathbf{o}$ and direction $\mathbf{d}_j$, the depth of a candidate voxel primitive centered at $\mathbf{p}_i$ is
\begin{equation}
t_{ij}
=
\mathbf{d}_j^{\mathsf T}(\mathbf{p}_i-\mathbf{o}).
\label{eq:ray_surface_depth}
\end{equation}
The candidates are examined in ascending order of $t_{ij}$. Let $r_i$ denote the spherical-support radius of the corresponding occupancy voxel. The normalized transverse distance from the ray to the voxel center is
\begin{equation}
m_{ij}
=
\frac{
\left\|
(\mathbf{I}-\mathbf{d}_j\mathbf{d}_j^{\mathsf T})
(\mathbf{p}_i-\mathbf{o})
\right\|_2^2
}{r_i^2},
\qquad
m_{ij}\leq 1,
\label{eq:ray_surface_transverse_distance}
\end{equation}
which tests whether the ray intersects the spherical support of the voxel. Candidates outside the near--far range or failing this test are rejected. The first valid candidate is returned as the visible surface; rays without a valid candidate are invalid. The rasterizer outputs a surface index and depth for each ray (Fig.~\ref{fig:ray-traced-rasterizer}).

\subsection{Ray-Based View Visibility}
\label{sec:ray-based-view-visibility}

\begin{figure}[!t]
  \centering
  \includegraphics[width=\columnwidth]{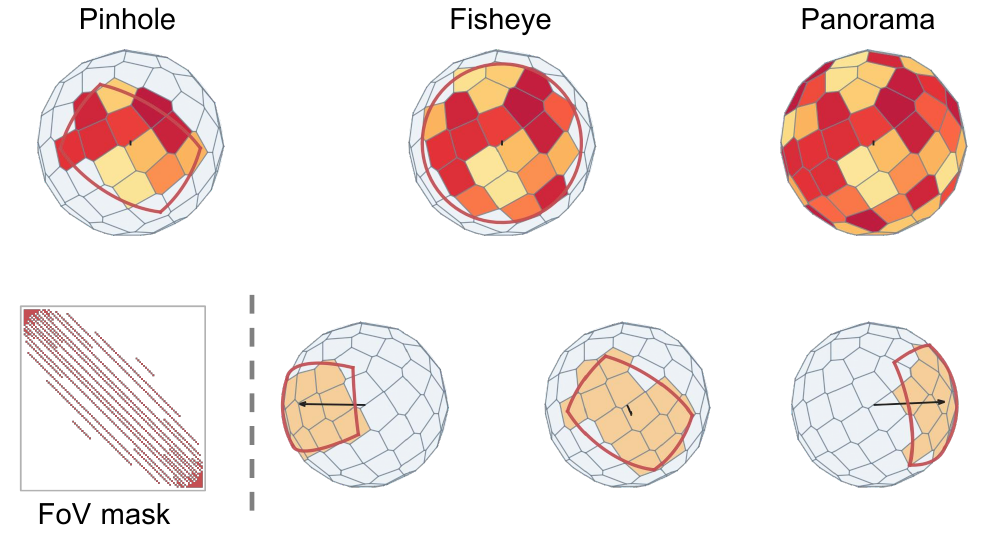}
  \caption{Ray-based view visibility for pinhole, fisheye, and panoramic RGB cameras. Dense camera rays are quantized into Fibonacci direction bins to form a camera-frame spherical footprint. For each candidate orientation, we express the spherical bin directions in the camera frame and test them against this footprint. The resulting mask identifies the bins used for gain aggregation.}
  \label{fig:ray-based-view-visibility}
\end{figure}

In the planning stage, R3CON represents the omnidirectional gain of a candidate view on the unit sphere and discretizes the sphere into direction bins using uniform Fibonacci sampling~\cite{r3con}. Each direction bin represents a group of nearby observation directions and stores the corresponding renderability gain. R3CON precomputes a field-of-view (FoV) mask between candidate view directions and direction bins:
\begin{equation}
M_i(j)\in\{0,1\},
\label{eq:fov-mask}
\end{equation}
where $M_i(j)=1$ indicates that direction bin $j$ is visible from candidate view $i$. Row $i$ of the mask identifies the bins within the camera's FoV at candidate orientation $i$, as illustrated in Fig.~\ref{fig:ray-based-view-visibility}. The directional gain of the candidate view is then obtained by aggregating the visible bins:
\begin{equation}
G_i=\sum_j M_i(j)g(j),
\label{eq:fov-gain-aggregation}
\end{equation}
where $g(j)$ denotes the renderability gain of direction bin $j$.

The accuracy of the FoV mask directly affects candidate-view gain evaluation. A mask constructed from fixed horizontal and vertical field-of-view angles implicitly assumes a regular rectangular boundary on the viewing sphere. This assumption is inadequate for fisheye and panoramic cameras, whose actual visible regions may have curved boundaries or cover a substantially larger portion of the sphere. The different FoV-mask footprints induced by pinhole, fisheye, and panoramic cameras are shown in Fig.~\ref{fig:ray-based-view-visibility}.

For camera model $\mathcal{C}$, its dense rays are assigned to the spherical direction bins to obtain a canonical visible-direction footprint in the camera frame:
\begin{equation}
\mathcal{S}_{\mathcal{C}}
=
\left\{
\operatorname{bin}(\mathbf{d}_p)
\mid
\mathbf{d}_p\in\mathcal{D}_{\mathcal{C}}
\right\},
\label{eq:camera_ray_footprint}
\end{equation}
where $\mathcal{D}_{\mathcal{C}}$ denotes the dense ray set generated by camera model $\mathcal{C}$ and $\operatorname{bin}(\cdot)$ maps a ray direction to its spherical direction bin. Because the dense rays are generated by the camera projection, this footprint encodes the visible directions of that camera model in its own coordinate frame; the outlines in Fig.~\ref{fig:ray-based-view-visibility} visualize these footprints.

For candidate view $i$, each spherical direction $\mathbf{q}_j$ is transformed into the camera frame associated with that view, and its visibility is determined by querying the canonical footprint:
\begin{equation}
M_i(j)
=
\mathbb{I}
\left[
\operatorname{bin}
\left(
\mathbf{R}_i^{\mathsf T}\mathbf{q}_j
\right)
\in
\mathcal{S}_{\mathcal{C}}
\right],
\label{eq:ray-based-mask}
\end{equation}
where $\mathbf{q}_j$ is the spherical direction associated with bin $j$, $\mathbf{R}_i$ denotes the orientation of candidate view $i$, and $\mathbb{I}[\cdot]$ is the indicator function. The rows index candidate viewing orientations, and the columns index the shared spherical direction bins. Each camera model specifies its visible direction-bin set through its canonical footprint, while mask queries and gain aggregation operate in the shared bin domain.

\section{Hierarchical Active Capture Planning}
\subsection{Global Region Management}
\label{sec:global-region-management}

\begin{figure}[!t]
    \centering
    \includegraphics[width=\columnwidth]{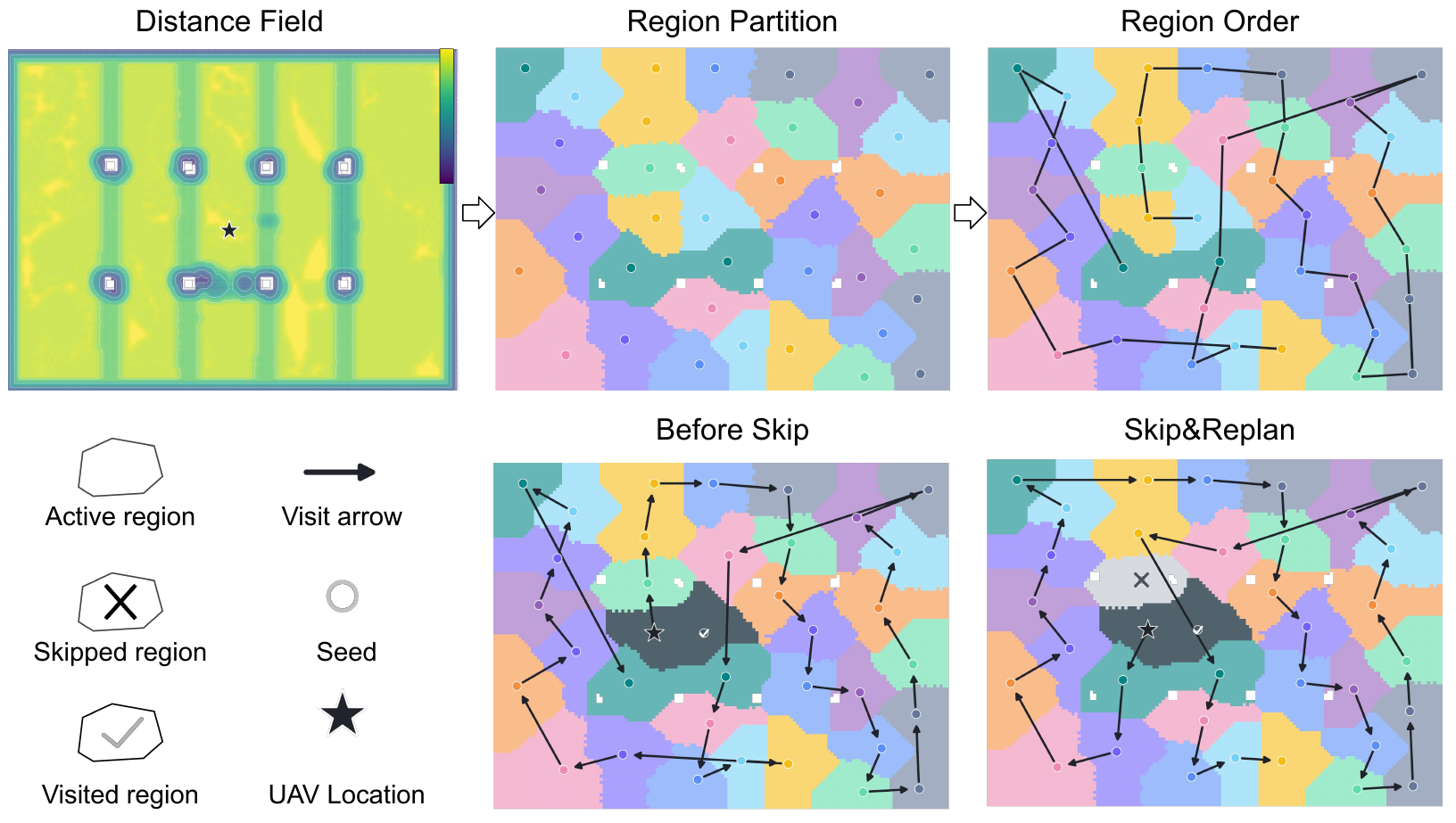}
    \caption{Global region management for large-scale active capture. The upper panels show the free-space distance field, seed-centered region partition, and initial visit order. In the lower panels, completed regions are dark gray and check-marked, the UAV is indicated by a star, and arrows show the remaining order. A region that fails the reconstruction-frontier test is cross-marked and removed before the remaining regions are reordered from the current UAV position.}
    \label{fig:global-region-management}
\end{figure}

Large scenes contain non-uniform free space and reconstructable surfaces. Searching the entire scene for every NBV expands the candidate set and increases cross-region motion. We partition reachable free space into bounded regions, each represented by a seed, to provide a controlled search scope for local NBV planning, as illustrated in Fig.~\ref{fig:global-region-management}.

From the free-space map provided by geometric exploration, we retain voxels that satisfy the clearance constraint and are connected to the UAV's current position. Local maxima of the free-space distance field form seed candidates, and a spatial-separation constraint selects the final seeds. Each seed serves as a query location for its region.

We compute seed-to-voxel connectivity distances on the reachable free-space graph and assign each voxel to its nearest seed. Compared with Euclidean nearest-seed assignment, graph-distance partitioning follows free-space connectivity across obstacles and narrow passages. Connectivity distances from the current UAV position define the initial visit order.

We determine task completion from the reconstruction gain expected from further acquisition. To estimate the remaining gain online, we query sampled positions in the current region and the seeds of regions awaiting a visit. At each queried position $p_i$, we compute the renderability gains for the surfaces visible from that position:

\begin{equation}
    \mathcal{G}(p_i) = \{g_{i,1}, g_{i,2}, \ldots, g_{i,N_i}\}.
\end{equation}

The residual value at $p_i$ is the 90th percentile of these gains:

\begin{equation}
    \rho(p_i) = Q_{\alpha}\big(\mathcal{G}(p_i)\big),
\end{equation}

where $Q_{\alpha}(\cdot)$ denotes the $\alpha$-quantile and $\alpha=0.9$. Given the reconstruction-value threshold $\tau_{\mathrm{fine}}$, a position is defined as a \emph{reconstruction-frontier viewpoint} if

\begin{equation}
    \rho(p_i) \geq \tau_{\mathrm{fine}}.
\end{equation}

For region $r$, the seed value is

\begin{equation}
    \rho_r^{\mathrm{seed}} = \rho(s_r).
\end{equation}

We retain region $r$ in the visit queue if $\rho_r^{\mathrm{seed}}\geq\tau_{\mathrm{fine}}$ and remove it otherwise.

For a region selected from the visit queue, let $\mathcal{P}_r$ denote its candidate positions. The candidate positions satisfying the residual-value criterion are

\begin{equation}
    \mathcal{F}_r =
    \left\{p_i \in \mathcal{P}_r \mid
    \rho(p_i) \geq \tau_{\mathrm{fine}}\right\}.
\end{equation}

If $\mathcal{F}_r$ is non-empty, Regional NBV Planning selects an executable capture viewpoint from this set. Otherwise, the system completes processing of the current region and continues with the next region in the visit queue.

The global scheduler maintains the visit order of retained regions. Completed and skipped regions are removed from the order; the remaining regions are reordered by inter-region motion cost from the current UAV position or current region. This order determines the region scope for local NBV planning, while Regional NBV Planning selects the position and direction within that scope.

Capture terminates when the visit queue is empty.

\subsection{Regional NBV Planning}
\label{sec:regional-nbv-planning}

Within the selected region, Regional NBV Planning evaluates sampled position--direction pairs according to their spatial renderability gains and effective time costs. We define the next-best view (NBV) as the executable pair with the largest spatial renderability gain per unit effective time among positions that meet the reconstruction-frontier criterion.

\subsubsection{Candidate View Generation}
\label{sec:candidate-view-generation}

The global region manager provides a region label for each reachable free voxel. For the currently selected region $r$, we retain the free-space locations that satisfy the safety-clearance constraint:

\begin{equation}
    \mathcal{F}^{\mathrm{safe}}_r
    =
    \left\{
    p \mid \operatorname{label}(p)=r,\;
    p\in\mathcal{F},\;
    d_{\mathrm{ESDF}}(p)\geq d_{\min}
    \right\},
\end{equation}

where $\mathcal{F}$ is reachable free space, $d_{\mathrm{ESDF}}(p)$ is the distance to the nearest occupied cell, and $d_{\min}$ is the safety clearance. We use farthest-point sampling (FPS) to select $K$ spatially separated positions that cover feasible observation locations in the current region:

\begin{equation}
    p_{k+1}
    =
    \underset{p\in\mathcal{F}^{\mathrm{safe}}_r}{\arg\max}
    \;\min_{p_j\in\mathcal{P}_k}
    \lVert p-p_j\rVert_2,
\end{equation}

where $\mathcal{P}_k$ is the set selected after $k$ iterations. The resulting candidate set is $\mathcal{P}_r=\{p_1,\ldots,p_K\}$. This step generates candidate positions; their directional gains and motion costs are evaluated subsequently.

\subsubsection{Spatial Renderability Gain}
\label{sec:spatial-renderability-gain}

For candidate position $p_i$ and candidate viewing direction $j$, we approximate the spatial renderability gain as

\begin{equation}
    G_i(j)
    \approx
    \int_{\mathcal{S}_{i,j}}
    u(\mathbf{x})\,\mathrm{d}A,
\end{equation}

where $\mathcal{S}_{i,j}$ is the visible surface, $u(\mathbf{x})$ is renderability gain at surface location $\mathbf{x}$, and $\mathrm{d}A$ denotes surface area.

Pre-sampled rays query visible surface points. Because an angular bin subtends a larger surface area at greater depth, summing ray gains directly would overrepresent nearby surfaces. Let $u_{i,n}$ and $z_{i,n}$ denote the gain and depth returned by ray $n$. We therefore weight each contribution by its represented surface area:

\begin{equation}
    \widetilde{u}_{i,n}
    =
    u_{i,n}\,a(z_{i,n}),
    \qquad
    a(z_{i,n})
    =
    \max\left(
    z_{i,n}^{2}\frac{\Omega_{\mathrm{bin}}}{s_v^{2}},1
    \right),
\end{equation}

where $\Omega_{\mathrm{bin}}$ is the direction-bin solid angle and $s_v$ is the voxel size. The term $z_{i,n}^{2}\Omega_{\mathrm{bin}}$ approximates the surface area subtended by the direction bin at depth $z_{i,n}$. Dividing by $s_v^2$ expresses this area in units of voxel face area. We clamp the resulting weight to at least one. Weighted contributions are accumulated into spherical direction bins:

\begin{equation}
    A_i(b)
    =
    \sum_{n:\,b(n)=b}
    \widetilde{u}_{i,n},
\end{equation}

where $b(n)$ is the bin of ray $n$. The precomputed FoV mask $M_{\mathrm{fov}}(j,b)$ associated with candidate viewing direction $j$ then yields its gain:

\begin{equation}
    G_i(j)
    =
    \sum_b M_{\mathrm{fov}}(j,b)A_i(b).
\end{equation}

The gains for all candidate positions and viewing directions form the position--direction gain matrix

\begin{equation}
    \mathbf{G}=[G_i(j)]\in\mathbb{R}^{K\times N_d},
\end{equation}

where $N_d$ is the number of candidate viewing directions.

\subsubsection{Dynamics-Aware NBV Selection}
\label{sec:dynamics-aware-nbv-selection}

For each candidate position--direction pair, we estimate a motion cost for NBV ranking. Let $\pi_i$ denote the path from the current UAV state to candidate position $p_i$. Its translational cost is modeled as

\begin{equation}
    T_i^{\mathrm{trans}}
    =
    \mathcal{T}_{v_{\max},a_{\max}}(\pi_i,\mathbf{v}_t),
\end{equation}

where $v_{\max}$ and $a_{\max}$ are the translational velocity and acceleration limits, and $\mathbf{v}_t$ is the current translational velocity. For a current direction $k$ and target direction $j$, the yaw change is $\Delta\psi_{k,j}$. The yaw-cost surrogate assumes zero initial yaw rate and does not impose a terminal yaw-rate constraint; under bounded angular velocity and angular acceleration, it is

\begin{equation}
    \begin{aligned}
    T_{k,j}^{\mathrm{yaw}}
    &=
    \begin{cases}
    \sqrt{\dfrac{2|\Delta\psi_{k,j}|}{\alpha_{\max}}},
    & |\Delta\psi_{k,j}|\leq\Delta\psi_{\mathrm{sw}},\\[8pt]
    \dfrac{\omega_{\max}}{\alpha_{\max}}
    +\dfrac{|\Delta\psi_{k,j}|-\Delta\psi_{\mathrm{sw}}}{\omega_{\max}},
    & |\Delta\psi_{k,j}|>\Delta\psi_{\mathrm{sw}},
    \end{cases}\\
    \Delta\psi_{\mathrm{sw}}
    &=\frac{\omega_{\max}^{2}}{2\alpha_{\max}},
    \end{aligned}
\end{equation}

where $\omega_{\max}$ and $\alpha_{\max}$ are the angular velocity and angular acceleration limits. This cost is used for NBV ranking; the subsequent trajectory planner generates the executed yaw trajectory. Since translation and yaw rotation can be executed in parallel, the ranking cost for a position--direction pair is

\begin{equation}
    T_{i,j}^{\mathrm{motion}}
    =
    \max\left(
    T_i^{\mathrm{trans}},T_{k,j}^{\mathrm{yaw}}
    \right).
\end{equation}

We precompute a pitch-validity mask on the discrete direction bins. Let $k$ denote the current direction under nominal body pitch, and let $\Delta\theta_{k,j}$ measure the pitch offset to candidate bin $j$. Because an exact zero-pitch bin need not exist, we use a binning tolerance of $\epsilon_\theta=2^\circ$:

\begin{equation}
    M_{k,j}^{\mathrm{pitch}}
    =
    \mathbf{1}\left[
    |\Delta\theta_{k,j}|\leq\theta_{\max}+\epsilon_\theta
    \right].
\end{equation}

Here, $\theta_{\max}$ is the allowable optical-axis pitch offset determined by the camera mounting configuration; executed NBVs retain the nominal body pitch. To account for the computation required by each closed-loop replanning step, we impose the planning time $T_{\mathrm{plan}}$ as a lower bound on the effective time cost:

\begin{equation}
    \overline{T}_{i,j}
    =
    \max\left(
    T_{i,j}^{\mathrm{motion}},T_{\mathrm{plan}}
    \right).
\end{equation}

When the estimated motion time is shorter than the estimated planning time, $T_{\mathrm{plan}}$ determines the effective cost.

Let $\mathcal{C}_r$ contain the candidate pairs $(i,j)$ with $p_i\in\mathcal{F}_r$, $M_{k,j}^{\mathrm{pitch}}=1$, and finite motion cost, where $\mathcal{F}_r$ is the set of reconstruction-frontier viewpoints in the current region. We select the next-best view from this set:

\begin{equation}
    (i^\star,j^\star)
    =
    \underset{(i,j)\in\mathcal{C}_r}{\arg\max}\;
    \frac{G_i(j)}{\overline{T}_{i,j}}.
\end{equation}

The planner returns the selected position $p_{i^\star}$ and target yaw $\psi_{j^\star}$ to the exploration and trajectory-execution modules.

\subsection{Continuous Event-Driven Capture}
\label{sec:continuous-event-driven-capture}

\begin{figure}[!t]
  \centering
  \includegraphics[width=\columnwidth]{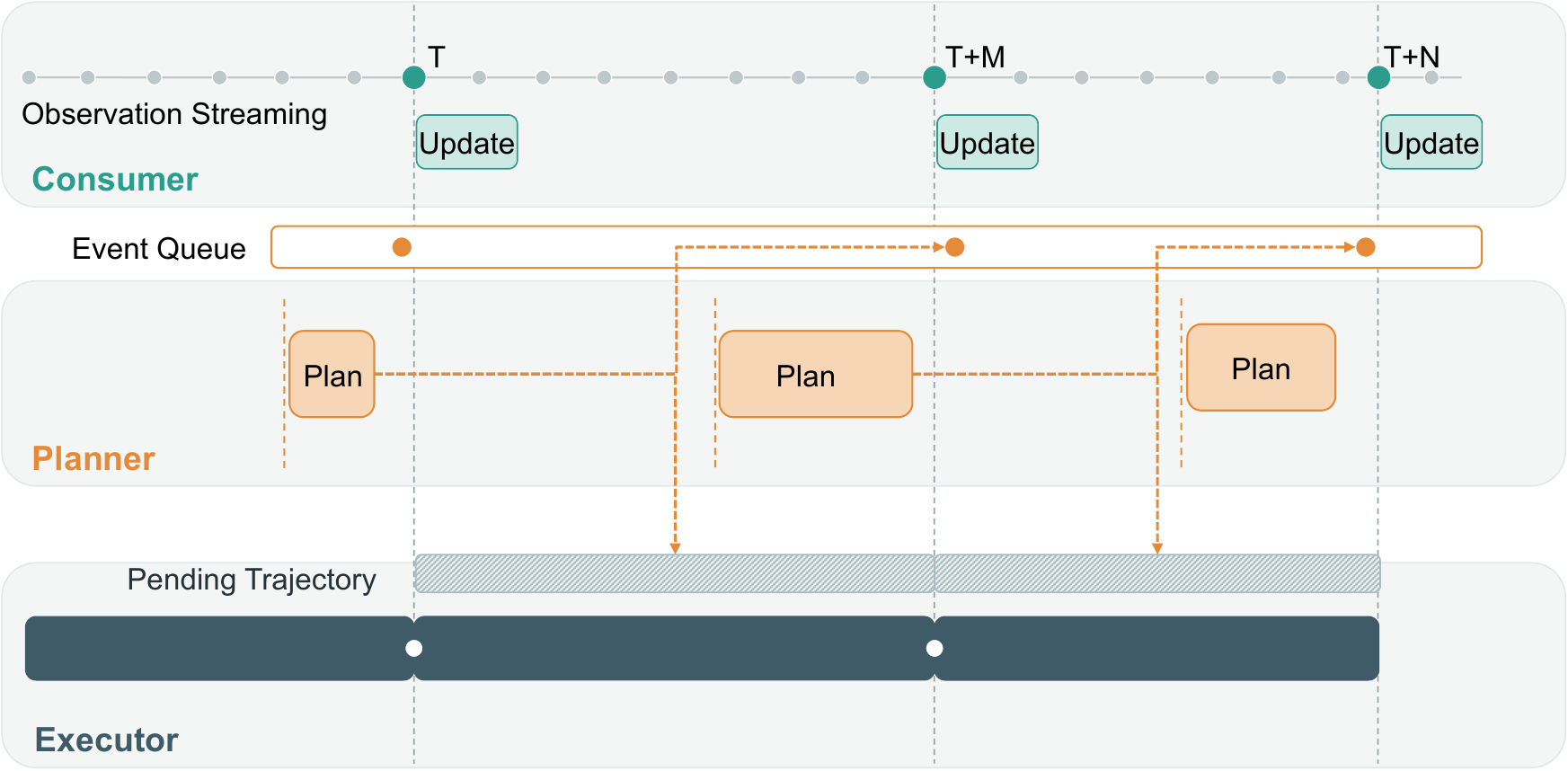}
  \caption{Asynchronous event-driven capture across consumer, planner, and executor timelines. Timestamped target events are queued and matched to the observation stream by the consumer. Planning overlaps with flight, and the next trajectory is queued for execution from the active trajectory's terminal state.}
  \label{fig:event-driven-capture}
\end{figure}

As shown in Fig.~\ref{fig:event-driven-capture}, continuous event-driven capture operates the consumer, planner, and executor asynchronously: the consumer processes observation events, the planner generates subsequent decisions while the current trajectory is executed, and the executor follows successive trajectory segments.

\subsubsection{Timestamped Event-Driven Observation}
\label{sec:timestamped-event-driven-observation}

During exploration, timestamps are published at \(1\,\mathrm{Hz}\) to record periodic observations. During reconstruction, the planner uses the time-parameterized trajectory to predict when the UAV will reach the selected NBV and publishes the corresponding timestamp to the event queue. The update front end continuously matches incoming synchronized RGB--pose data to queued timestamps; only matched observations update the persistent renderability state.

\subsubsection{Early Replanning}
\label{sec:early-replanning}

Early replanning starts the next planning step before the current trajectory ends. A fixed-length first-in, first-out queue \(\mathcal{H}\) stores recent planning runtimes. It is initialized with \(T_0\); after each planning step, the measured runtime \(\hat{T}_{\mathrm{plan}}^{(n)}\) is appended. The planning-time estimate is the queue maximum scaled by \(\gamma\):
\begin{equation}
T_{\mathrm{plan}}^{(n)}
=
\gamma
\max_{\hat{T}\in\mathcal{H}^{(n)}}\hat{T},
\label{eq:planning-time-estimate}
\end{equation}

where \(\mathcal{H}^{(n)}\) is the queue after the \(n\)-th planning step. We use a queue length of \(5\) and \(\gamma=1.2\).

Let \(T_{\mathrm{remain}}\) denote the remaining time of the currently executed trajectory. The system initiates replanning when \(T_{\mathrm{remain}}\leq T_{\mathrm{plan}}^{(n)}\). The scaled maximum of recent planning runtimes sets the trigger threshold so that replanning can overlap with execution.

\subsubsection{Reversible Pseudo-Update and Trajectory Stitching}
\label{sec:reversible-pseudo-update}

During early replanning, the preceding NBV may not yet have produced a real observation. The \emph{persistent renderability state} reflects only timestamp-matched observations; replanning directly from this state can therefore select another view of the same surfaces. To form the \emph{planning state}, we predict visible surface points, ray directions, and ray distances from the preceding target pose and temporarily update the affected entries, saving their previous values.

Following the renderability-state definition in~\eqref{eq:renderability-state}, let \(g\) denote the preceding NBV target and \(\widehat{\mathcal{S}}_g\) the visible surface set predicted from its target pose. The pseudo-updated planning state is
\begin{equation}
    \widehat{\mathcal{R}}_x^{+}
    =
    \begin{cases}
    \left(
    \mathcal{A}_x,\;
    \mathcal{D}_x\cup\{\widehat{\mathbf{r}}_{g,x}\},\;
    \min(d_x,\widehat{d}_{g,x})
    \right),
    & x\in\widehat{\mathcal{S}}_g,\\
    \mathcal{R}_x,
    & x\notin\widehat{\mathcal{S}}_g,
    \end{cases}
    \label{eq:pseudo-renderability-update}
\end{equation}
where \(\widehat{\mathbf{r}}_{g,x}\) and \(\widehat{d}_{g,x}\) are the predicted ray direction and distance from the target pose to \(x\). The pseudo-update preserves appearance statistics while accounting for the predicted direction coverage and observation distance. Candidate queries on this state account for surfaces expected to be observed at the preceding NBV.

After querying the candidates, the system restores the saved entries and uses the query results both to select the next NBV and to assess the remaining reconstruction gain in the current region. When the target event arrives, the update front end consumes the timestamp-matched RGB--pose observation and updates the persistent renderability state.

The planner generates the subsequent time-parameterized trajectory from the terminal state of the active trajectory and stitches it for execution. The subsequent trajectory is planned during execution of the active trajectory and committed before it ends, maintaining continuous motion.

\section{Experiments}
\label{sec:experiments}

\subsection{Experimental Setup and Evaluation Protocol}
\label{sec:experimental-setup}

\begin{table*}[!t]
  \centering
  \caption{Comparison protocols for online active capture and offline view placement.}
  \label{tab:experiment_protocol}
  \small
  \setlength{\tabcolsep}{4.5pt}
  \renewcommand{\arraystretch}{1.08}
  \begin{tabularx}{\textwidth}{@{}>{\raggedright\arraybackslash}p{0.16\textwidth}>{\raggedright\arraybackslash}X>{\raggedright\arraybackslash}X@{}}
    \toprule
    \textbf{Protocol item} & \textbf{Online active capture} & \textbf{Offline view placement} \\
    \midrule
    \textbf{Methods}
    & Ours; ActiveGS; TRACE
    & Ours; PCP; FPS+Rand. \\
    \textbf{Backend}
    & Incremental 2DGS
    & Offline 3DGS \\
    \addlinespace[2pt]
    \textbf{Budget}
    & $900\,\mathrm{s}$ elapsed system time
    & Matched $N_s$ views per scene \\
    \textbf{Stopping}
    & Ours: completion or $900\,\mathrm{s}$; baselines: $900\,\mathrm{s}$
    & Exactly $N_s$ views \\
    \addlinespace[2pt]
    \textbf{Shared setup}
    & Initial pose, sensing configuration, planning space, and UAV dynamics
    & Scene, sensing configuration, camera parameters, and evaluation views \\
    \textbf{Hardware}
    & Ours: RTX 4060 Laptop GPU (8\,GB); baselines: A100 SXM4 GPU (64\,GB)
    & A100 SXM4 GPU (64\,GB) \\
    \bottomrule
  \end{tabularx}
\end{table*}

\begin{figure*}[!t]
  \centering
  \includegraphics[width=\textwidth]{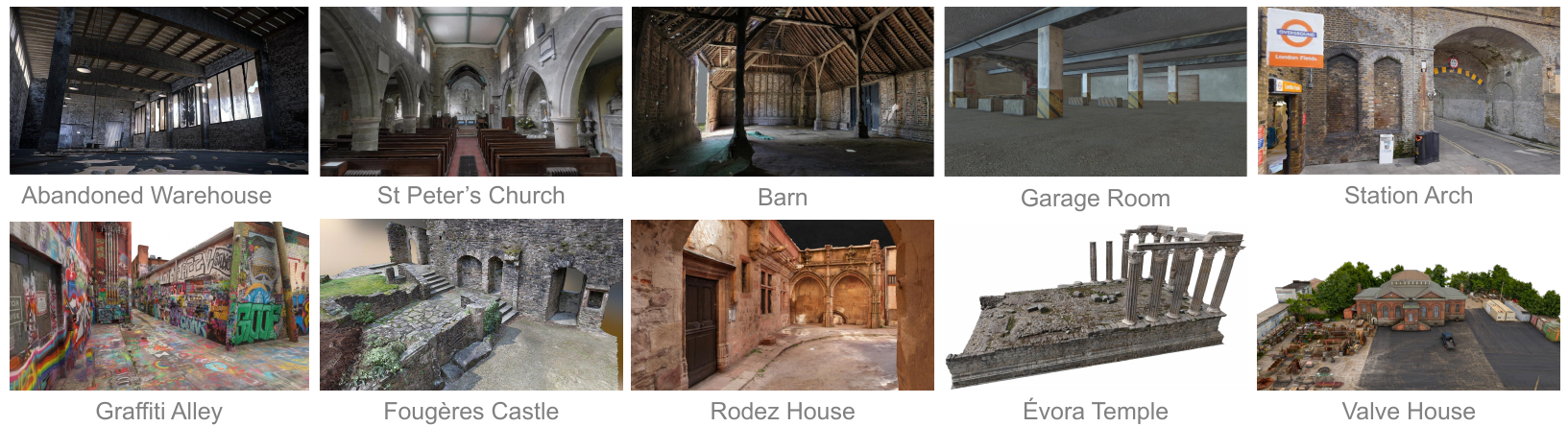}
  \caption{Ten Habitat scenes used in the main comparison, covering indoor and outdoor environments with different spatial scales, visibility conditions, and structural complexity.}
  \label{fig:benchmark-scenes}
\end{figure*}

In Habitat~\cite{savva2019habitat}, we evaluate online capture under a common mission-time cap and offline view placement under matched view counts. We compare AGE-Capture with two online baselines: ActiveGS~\cite{activegs2025}, a discrete NBV method, and TRACE~\cite{zheng2026trace}, a trajectory-level method. Offline comparisons include the view set acquired by our online system, Progressive Camera Placement (PCP)~\cite{kopanas2023improving}, and FPS+Random. As summarized in Table~\ref{tab:experiment_protocol}, online methods use a common incremental backend based on 2D Gaussian Splatting (2DGS)~\cite{huang2024_2dgs}, whereas offline methods train standard 3D Gaussian Splatting (3DGS) from complete view sets.

The main comparison uses a $640\times480$ pinhole RGB-D sensor with $f_x=f_y=320$ and a $0.1$--$5.0\,\mathrm{m}$ depth range. All online methods share the initial UAV pose, sensing configuration, planning space, and translational and yaw constraints; they control camera $x$, $y$, $z$, and yaw while maintaining nominal body pitch and roll.

The online time budget is $900\,\mathrm{s}$. For discrete NBV methods, motion, map updates, and planning occur sequentially:
\begin{equation}
T_{\mathrm{sys}}^{\mathrm{NBV}}
=
\sum_k
\left(
T_{\mathrm{motion}}^{(k)}
+
T_{\mathrm{mapping}}^{(k)}
+
T_{\mathrm{planning}}^{(k)}
\right).
\label{eq:nbv-system-time}
\end{equation}
Our asynchronous event-driven system uses the elapsed time from mission start to termination,
\begin{equation}
T_{\mathrm{sys}}^{\mathrm{ours}}
= t_{\mathrm{end}}-t_{\mathrm{start}}.
\label{eq:ours-system-time}
\end{equation}
where overlapping planning and execution intervals are counted once. All methods use the same bounded-dynamics model for translational and yaw motion. For TRACE, we approximate the path length by summing the distances between consecutive waypoints and use the same motion model to estimate execution time. Habitat rendering, sensor queries, and interface calls are excluded from the reported system time. Baselines run to $900\,\mathrm{s}$; ours stops at reconstruction-driven completion or the time limit.

The offline experiment isolates the effect of view placement on reconstruction quality. For each scene, the number of timestamp-matched RGB--pose observations that our online system successfully writes to the reconstruction backend before termination defines the view budget $N_s$. This value is fixed as the common budget for all offline methods. Each method generates exactly $N_s$ views under identical scene and camera settings and trains standard 3DGS using the complete view set.

The main comparison covers ten indoor and outdoor scenes (Fig.~\ref{fig:benchmark-scenes}) sourced from Sketchfab\footnote{\urlstyle{same}\url{https://sketchfab.com/}}. Evaluation positions are sampled from $0.2\,\mathrm{m}$ free-space voxels by farthest-point sampling with a $2\,\mathrm{m}$ minimum separation. At each evaluation position, we render six candidate views: four horizontal views spaced $90^\circ$ apart and two views looking straight up and down. We discard views in which fewer than $20\%$ of pixels have valid depth.

We report PSNR, structural similarity (SSIM)~\cite{wang2004image}, and learned perceptual image patch similarity (LPIPS)~\cite{zhang2018unreasonable} on the common evaluation images, together with the system time $T_{\mathrm{sys}}$, the number $N_{\mathrm{event}}$ of timestamp-matched RGB--pose observations successfully written to the reconstruction backend, and peak GPU memory.
\begin{table}[!t]
\centering
\caption{Per-scene acquisition statistics of our online system. The count of timestamp-matched RGB--pose observations written to the reconstruction backend defines the matched offline view budget.}
\label{tab:acquisition-statistics}
\begin{tabularx}{\linewidth}{l>{\centering\arraybackslash}X >{\centering\arraybackslash}X}
\toprule
Scene & Matched Observations & System Time (s) \\
\midrule
Abandoned Warehouse & 339 & 461.0 \\
St Peter's Church & 520 & 760.7 \\
Barn & 393 & 536.5 \\
Garage Room & 552 & 791.7 \\
Station Arch & 321 & 430.0 \\
Graffiti Alley & 593 & 837.2 \\
Fougères Castle & 117 & 152.8 \\
Rodez House & 155 & 205.8 \\
Évora Temple & 442 & 592.3 \\
Valve House & 347 & 470.5 \\
\bottomrule
\end{tabularx}
\end{table}

\subsection{Reconstruction Quality}
\label{sec:reconstruction-quality}

\setcounter{dbltopnumber}{2}

\begin{table*}[!t]
\centering
\small
\setlength{\tabcolsep}{4pt}
\caption{Online reconstruction quality under the 900 s mission-time cap. Best and second-best values within each row and metric are \textbf{bold} and \underline{underlined}, respectively.}
\label{tab:online-reconstruction-quality}
\begin{tabularx}{\textwidth}{l>{\centering\arraybackslash}X >{\centering\arraybackslash}X >{\centering\arraybackslash}X >{\centering\arraybackslash}X >{\centering\arraybackslash}X >{\centering\arraybackslash}X >{\centering\arraybackslash}X >{\centering\arraybackslash}X >{\centering\arraybackslash}X}
\toprule
 & \multicolumn{3}{c}{PSNR $\uparrow$} & \multicolumn{3}{c}{SSIM $\uparrow$} & \multicolumn{3}{c}{LPIPS $\downarrow$} \\
\cmidrule(lr){2-4}\cmidrule(lr){5-7}\cmidrule(lr){8-10}
Scene & ActiveGS & TRACE & Ours & ActiveGS & TRACE & Ours & ActiveGS & TRACE & Ours \\
\midrule
Abandoned Warehouse & \underline{24.19} & 21.19 & \textbf{24.92} & \underline{0.794} & 0.581 & \textbf{0.803} & \textbf{0.267} & 0.419 & \underline{0.270} \\
St Peter's Church & 20.62 & \underline{24.02} & \textbf{26.00} & 0.686 & \underline{0.764} & \textbf{0.802} & 0.413 & \underline{0.341} & \textbf{0.327} \\
Barn & \underline{21.29} & 19.32 & \textbf{21.70} & \underline{0.660} & 0.620 & \textbf{0.672} & \textbf{0.319} & 0.368 & \underline{0.328} \\
Garage Room & \underline{24.53} & 23.53 & \textbf{26.07} & \underline{0.662} & 0.642 & \textbf{0.663} & \textbf{0.483} & \underline{0.489} & 0.546 \\
Station Arch & 15.48 & \underline{16.10} & \textbf{26.23} & 0.407 & \underline{0.419} & \textbf{0.756} & 0.644 & \underline{0.629} & \textbf{0.371} \\
Graffiti Alley & 17.47 & \underline{18.55} & \textbf{23.80} & 0.636 & \underline{0.667} & \textbf{0.748} & 0.388 & \underline{0.375} & \textbf{0.325} \\
Fougères Castle & 22.55 & \textbf{27.62} & \underline{26.42} & 0.664 & \textbf{0.752} & \underline{0.722} & 0.398 & \textbf{0.320} & \underline{0.359} \\
Rodez House & 25.07 & \underline{26.55} & \textbf{28.09} & 0.811 & \underline{0.822} & \textbf{0.836} & \underline{0.298} & \textbf{0.283} & 0.301 \\
Évora Temple & \underline{23.73} & 20.12 & \textbf{24.35} & \underline{0.748} & 0.719 & \textbf{0.754} & \textbf{0.264} & 0.287 & \underline{0.274} \\
Valve House & 25.04 & \underline{25.31} & \textbf{27.69} & \underline{0.747} & 0.736 & \textbf{0.792} & \underline{0.330} & 0.332 & \textbf{0.307} \\
\midrule
\textbf{Average} & 22.00 & \underline{22.23} & \textbf{25.53} & \underline{0.682} & 0.672 & \textbf{0.755} & \underline{0.380} & 0.384 & \textbf{0.341} \\
\bottomrule
\end{tabularx}
\end{table*}

\textbf{Online capture.} Our missions terminate after $152.8$--$837.2\,\mathrm{s}$, before the $900\,\mathrm{s}$ limit. At termination, AGE-Capture's mean PSNR exceeds that of ActiveGS by $3.53\,\mathrm{dB}$ and that of TRACE by $3.30\,\mathrm{dB}$, with corresponding SSIM and LPIPS improvements (Table~\ref{tab:online-reconstruction-quality}). Table~\ref{tab:acquisition-statistics} reports the corresponding timestamp-matched observation counts and system times. Mission termination times vary by scene because a mission ends only when no remaining region passes the reconstruction-frontier criterion.

\begin{figure*}[!t]
  \centering
  \includegraphics[width=\textwidth]{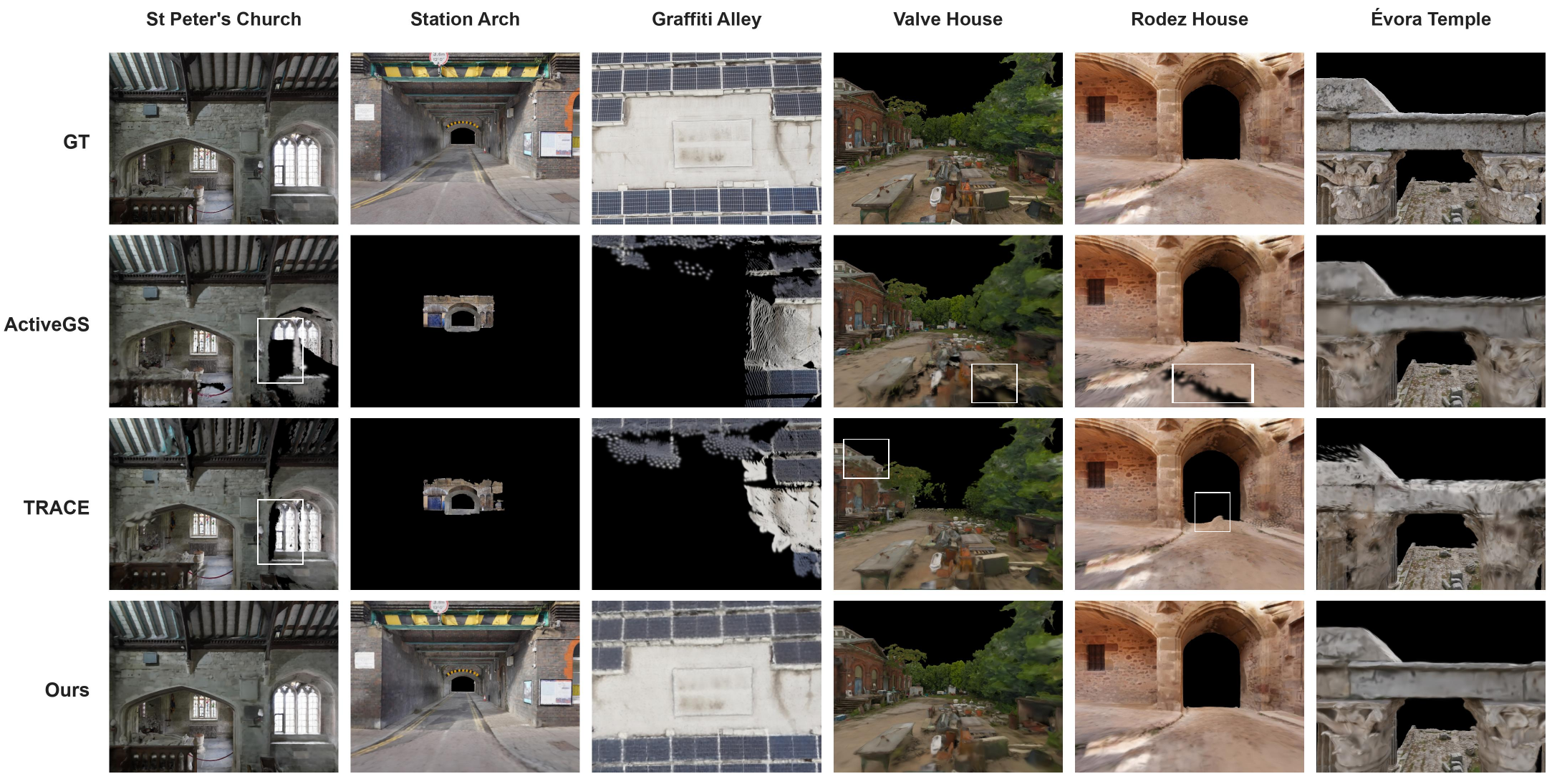}
  \caption{Qualitative online reconstruction under the common $900\,\mathrm{s}$ mission-time cap. Columns show six representative scenes; rows compare ground truth (GT), ActiveGS, TRACE, and ours. White boxes mark baseline regions with missing geometry or appearance degradation.}
  \label{fig:online-qualitative}
\end{figure*}

The largest online gains occur in scenes requiring broad spatial coverage. On Station Arch, PSNR exceeds the strongest baseline by $10.13\,\mathrm{dB}$. Figure~\ref{fig:online-qualitative} shows substantial missing geometry in the baseline reconstructions of both Station Arch and Graffiti Alley. In the baselines, reconstruction-oriented local planning can concentrate observations around local targets while leaving other reachable regions insufficiently explored. Separating geometric exploration from reconstruction-oriented capture maintains broad scene coverage while allocating additional observations to surfaces with residual reconstruction value.

\begin{table*}[t]
\centering
\small
\setlength{\tabcolsep}{4pt}
\caption{Offline reconstruction quality with matched per-scene view budgets. PCP denotes Progressive Camera Placement, and FPS denotes farthest-point sampling. Best and second-best values within each row and metric are \textbf{bold} and \underline{underlined}, respectively.}
\label{tab:offline-reconstruction-quality}
\begin{tabularx}{\textwidth}{l>{\centering\arraybackslash}X >{\centering\arraybackslash}X >{\centering\arraybackslash}X >{\centering\arraybackslash}X >{\centering\arraybackslash}X >{\centering\arraybackslash}X >{\centering\arraybackslash}X >{\centering\arraybackslash}X >{\centering\arraybackslash}X}
\toprule
 & \multicolumn{3}{c}{PSNR $\uparrow$} & \multicolumn{3}{c}{SSIM $\uparrow$} & \multicolumn{3}{c}{LPIPS $\downarrow$} \\
\cmidrule(lr){2-4}\cmidrule(lr){5-7}\cmidrule(lr){8-10}
Scene & {\scriptsize FPS+Random} & PCP & Ours & {\scriptsize FPS+Random} & PCP & Ours & {\scriptsize FPS+Random} & PCP & Ours \\
\midrule
Abandoned Warehouse & \underline{26.05} & 25.95 & \textbf{26.18} & \textbf{0.8387} & 0.8331 & \underline{0.8332} & \textbf{0.2108} & 0.2336 & \underline{0.2189} \\
St Peter's Church & \underline{24.41} & 21.93 & \textbf{25.22} & \underline{0.8109} & 0.7512 & \textbf{0.8315} & \underline{0.3402} & 0.4151 & \textbf{0.3164} \\
Barn & 20.40 & \underline{20.61} & \textbf{22.03} & \underline{0.6640} & 0.6637 & \textbf{0.7186} & \underline{0.3132} & 0.3324 & \textbf{0.2725} \\
Garage Room & \textbf{26.90} & 24.80 & \underline{26.84} & \textbf{0.7251} & 0.6783 & \underline{0.7166} & \textbf{0.4424} & 0.5174 & \underline{0.4597} \\
Station Arch & \underline{25.27} & 24.94 & \textbf{25.58} & \underline{0.7568} & 0.7413 & \textbf{0.7744} & \underline{0.3008} & 0.3140 & \textbf{0.2759} \\
Graffiti Alley & \underline{24.19} & 23.46 & \textbf{24.80} & \underline{0.7757} & 0.7423 & \textbf{0.7934} & \underline{0.2576} & 0.2975 & \textbf{0.2332} \\
Fougères Castle & 25.09 & \underline{25.35} & \textbf{25.38} & \textbf{0.6955} & 0.6563 & \underline{0.6763} & \textbf{0.2898} & 0.3388 & \underline{0.2932} \\
Rodez House & \textbf{26.99} & \underline{26.94} & 26.21 & \textbf{0.8105} & 0.7997 & \underline{0.8013} & \textbf{0.2588} & 0.2863 & \underline{0.2758} \\
Évora Temple & 24.43 & \underline{25.03} & \textbf{25.68} & \underline{0.7638} & 0.7506 & \textbf{0.7908} & \underline{0.2272} & 0.2431 & \textbf{0.2198} \\
Valve House & \underline{26.91} & 26.77 & \textbf{27.01} & \underline{0.7553} & 0.7278 & \textbf{0.7638} & \underline{0.2763} & 0.3003 & \textbf{0.2761} \\
\midrule
\textbf{Average} & \underline{25.06} & 24.58 & \textbf{25.49} & \underline{0.7596} & 0.7344 & \textbf{0.7700} & \underline{0.2917} & 0.3278 & \textbf{0.2842} \\
\bottomrule
\end{tabularx}
\end{table*}

\begin{figure*}[t]
  \centering
  \includegraphics[width=\textwidth]{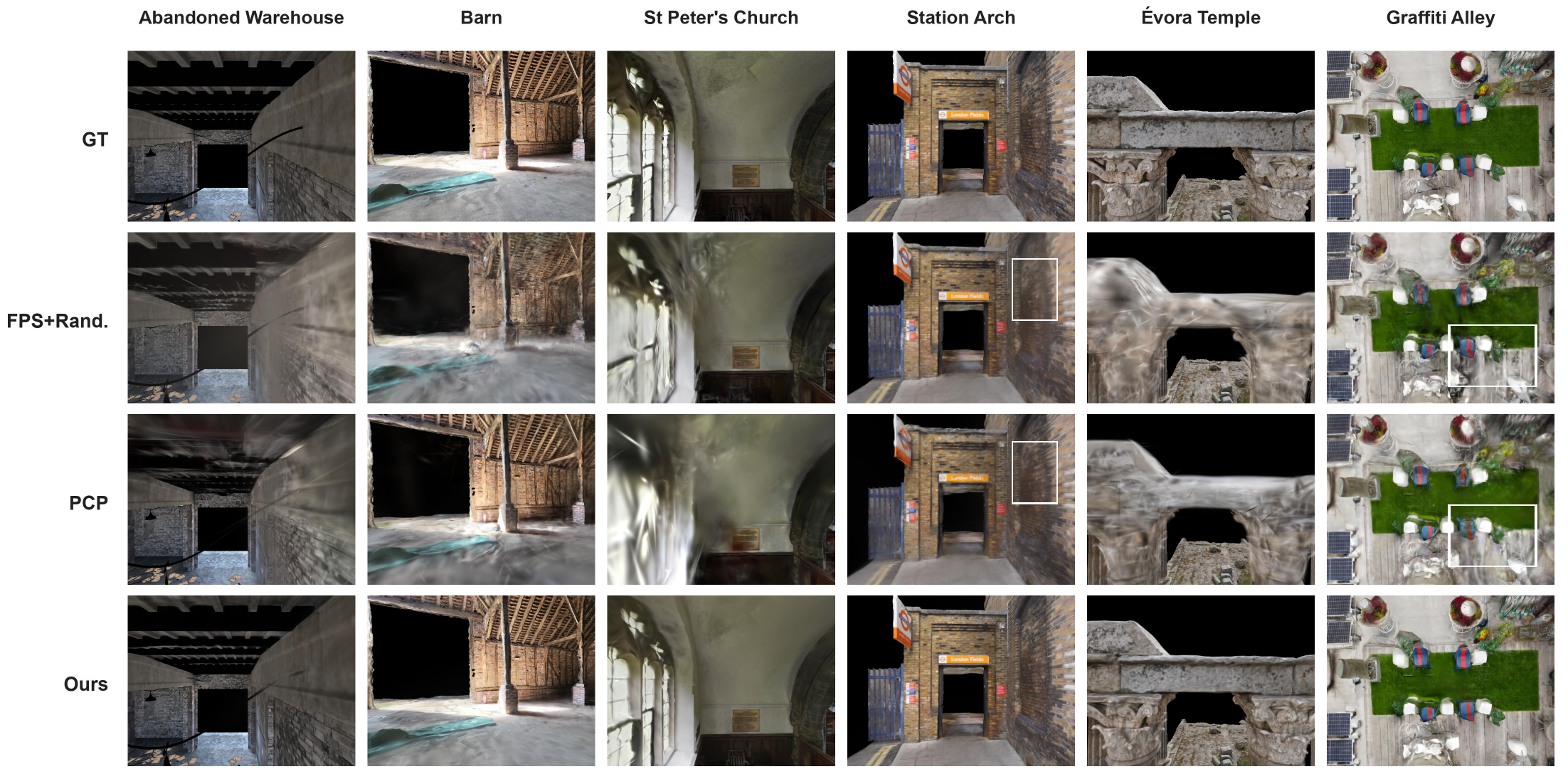}
  \caption{Qualitative offline reconstruction with matched per-scene view budgets. Rows compare ground truth (GT), FPS+Rand., Progressive Camera Placement (PCP), and ours across six scenes; white boxes mark representative baseline regions with structural or appearance degradation.}
  \label{fig:offline-qualitative}
\end{figure*}
\begin{figure*}[!t]
  \centering
  \includegraphics[width=\textwidth]{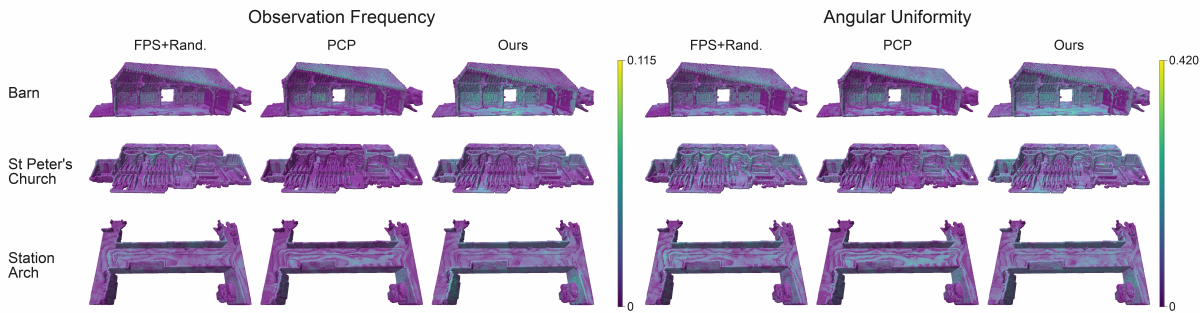}
  \caption{Surface observation support for equal-budget offline view placement. Observation frequency and angular uniformity are computed on a common surface point cloud by replaying selected poses with depth-occlusion tests. Brighter colors indicate stronger support; each metric uses a common scale across methods and scenes.}
  \label{fig:surface-observation-support}
\end{figure*}

\begin{figure*}[t]
  \centering
  \includegraphics[width=\textwidth]{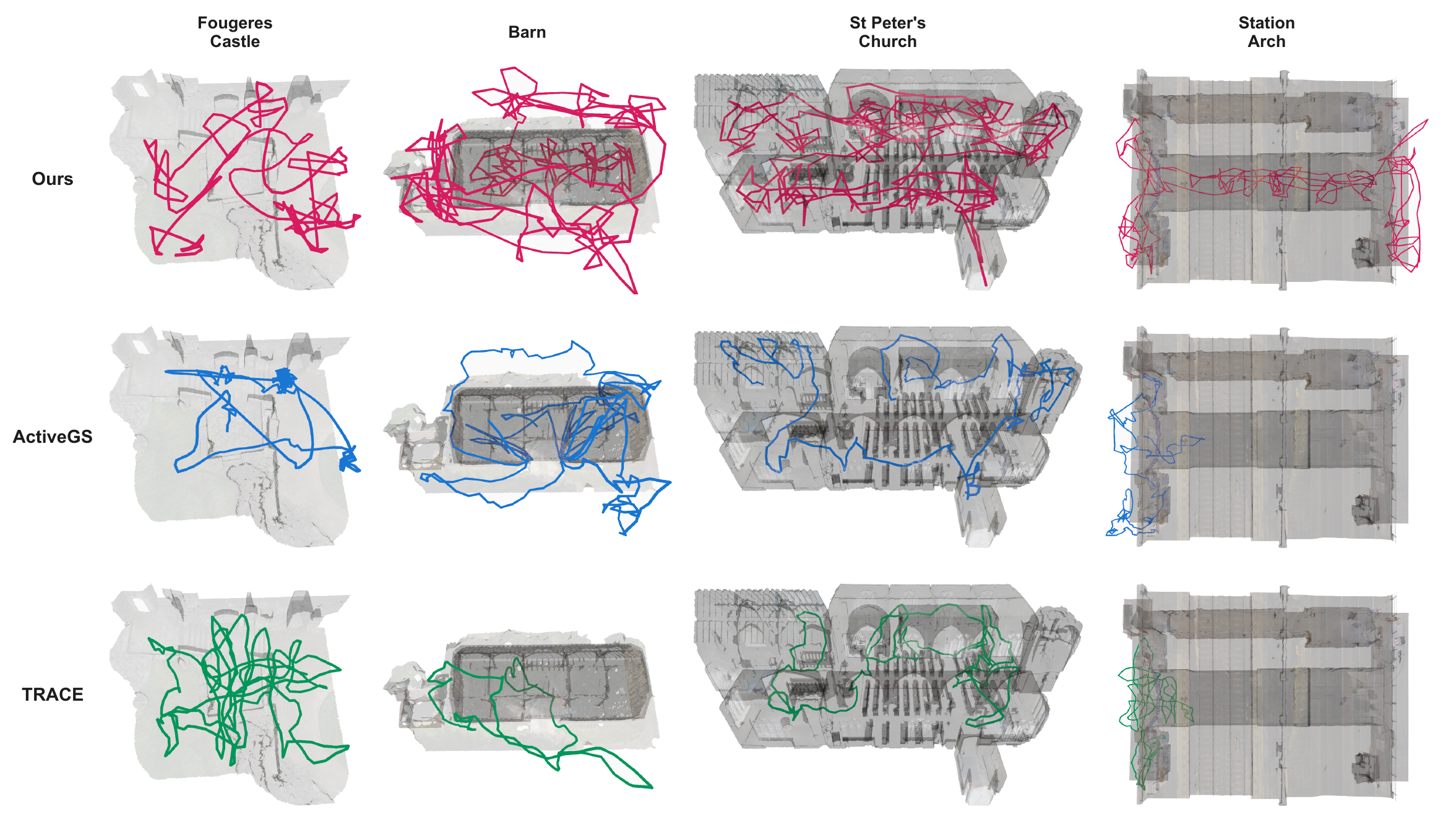}
  \vspace{1mm}
  \includegraphics[width=\textwidth]{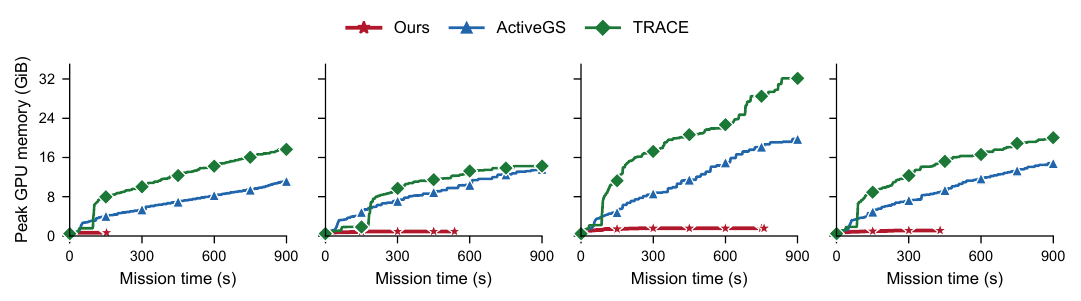}
  \caption{Execution trajectories and cumulative peak GPU memory for Foug\`eres Castle, Barn, St Peter's Church, and Station Arch, from left to right. The upper rows compare trajectories from ours, ActiveGS, and TRACE. The lower plots show cumulative peak GPU memory usage, with monitoring timestamps linearly mapped to each run's mission duration. Each trace ends at mission termination: the baselines run to $900\,\mathrm{s}$, whereas ours may terminate earlier under the reconstruction-driven completion criterion.}
  \label{fig:execution-efficiency}
\end{figure*}
\textbf{Offline view placement.} Under matched view budgets, FPS+Random distributes cameras throughout free space and achieves a mean PSNR within $0.43\,\mathrm{dB}$ of ours (Table~\ref{tab:offline-reconstruction-quality}), but its reconstructions show localized structural blur and appearance degradation (Fig.~\ref{fig:offline-qualitative}). This contrast shows that similar global average quality can still mask insufficient spatial or angular observation support in local regions. Its random orientations do not respond to the observation support already available at each surface. Renderability-guided selection jointly determines camera positions and orientations to supplement missing spatial and angular observation support.

To analyze these view-placement differences, we adapt two surface observation-support metrics from Progressive Camera Placement (PCP)~\cite{kopanas2023improving}: observation frequency and angular uniformity. Given $N_s$ cameras, let $v_i(p)$ indicate whether surface point $p$ lies within the field of view of camera $i$ and passes the depth-occlusion test. The two scalar fields are
\begin{equation}
F(p)=\frac{1}{N_s}\sum_{i=1}^{N_s}v_i(p),
\qquad
A(p)=1-\frac{1}{2}\sum_{b=1}^{B}
\left|h_p(b)-\frac{1}{B}\right|,
\label{eq:surface-observation-support}
\end{equation}
where $h_p(b)$ is the normalized histogram of valid viewing directions over $B$ angular bins. Figure~\ref{fig:surface-observation-support} shows that our method supplements observations in low-frequency regions and provides complementary viewing directions. FPS+Random produces irregular frequency and directional support, whereas PCP provides stronger observation support on exposed surfaces than in occluded regions. The low support on the Church floor and denser support inside Barn reflect the joint effects of occlusion and reachable free space. These support distributions explain the local reconstruction differences observed under the same view budget despite similar global average quality.

\begin{table*}[!t]
\centering
\small
\setlength{\tabcolsep}{4.5pt}
\renewcommand{\arraystretch}{1.04}
\caption{Sensor configurations used in the adaptability study.}
\label{tab:sensor-configurations}
\begin{tabularx}{\textwidth}{@{}>{\centering\arraybackslash}p{0.08\textwidth}>{\raggedright\arraybackslash}p{0.13\textwidth}>{\raggedright\arraybackslash}X>{\centering\arraybackslash}p{0.17\textwidth}>{\centering\arraybackslash}p{0.15\textwidth}>{\centering\arraybackslash}p{0.10\textwidth}@{}}
\toprule
\textbf{Stream} & \textbf{Configuration} & \textbf{Projection / model} & \textbf{Resolution / sampling} & \textbf{Angular coverage} & \textbf{Valid range} \\
\midrule
\raisebox{-0.5\baselineskip}{\textbf{Range}}
& Depth Map & Pinhole & $640\times480$ px & $90^\circ\times73.7^\circ$ & $0.1$--$5$ m \\
& LiDAR & 32-beam spinning & $32\times1024$ beams & $360^\circ\times45^\circ$ & $0.5$--$80$ m \\
\cmidrule(lr){1-6}
& Pinhole & Pinhole & $640\times480$ px & $90^\circ\times73.7^\circ$ & -- \\
\textbf{RGB} & Fisheye & Double sphere & $1280\times1280$ px & $160^\circ$ circular FoV & -- \\
& Panoramic & Equirectangular & $2880\times1440$ px & $360^\circ\times180^\circ$ & -- \\
\midrule
\multicolumn{6}{@{}l}{\footnotesize Pinhole: $f_x=f_y=320$ px. Double sphere: $f=458.37$ px, $\alpha=0.59$, and $\xi=-0.18$.} \\
\bottomrule
\end{tabularx}
\end{table*}

\begin{table*}[!t]
\centering
\scriptsize
\setlength{\tabcolsep}{2.1pt}
\caption{\MakeUppercase{Reconstruction Quality and Mission Time Across RGB Camera Models. Higher PSNR/SSIM, Lower LPIPS and Mission Time Are Better. Best and Second-Best Values within Each Row and Metric Are Bold and Underlined, Respectively.}}
\label{tab:rgb-camera-model-adaptability}
\begin{tabularx}{\textwidth}{>{\raggedright\arraybackslash}p{0.16\textwidth}*{12}{>{\centering\arraybackslash}X}}
\toprule
& \multicolumn{4}{c}{Pinhole} & \multicolumn{4}{c}{Fisheye} & \multicolumn{4}{c}{Panoramic} \\
\cmidrule(lr){2-5}\cmidrule(lr){6-9}\cmidrule(lr){10-13}
Scene & PSNR $\uparrow$ & SSIM $\uparrow$ & LPIPS $\downarrow$ & \shortstack{Mission\\Time (s) $\downarrow$} & PSNR $\uparrow$ & SSIM $\uparrow$ & LPIPS $\downarrow$ & \shortstack{Mission\\Time (s) $\downarrow$} & PSNR $\uparrow$ & SSIM $\uparrow$ & LPIPS $\downarrow$ & \shortstack{Mission\\Time (s) $\downarrow$} \\
\midrule
Valve House & 24.19 & 0.636 & 0.431 & 470.5 & \underline{26.51} & \textbf{0.731} & \underline{0.400} & \underline{449.5} & \textbf{27.04} & \underline{0.714} & \textbf{0.389} & \textbf{170.3} \\
St Peter's Church & 21.47 & 0.748 & 0.510 & 760.7 & \underline{24.17} & \underline{0.789} & \underline{0.453} & \underline{481.5} & \textbf{25.50} & \textbf{0.806} & \textbf{0.435} & \textbf{333.7} \\
Graffiti Alley & 20.88 & 0.636 & 0.419 & 837.2 & \underline{22.87} & \underline{0.716} & \underline{0.381} & \underline{502.0} & \textbf{23.84} & \textbf{0.736} & \textbf{0.363} & \textbf{341.3} \\
Fougères Castle & 23.37 & 0.606 & 0.484 & 168.3 & \underline{23.66} & \underline{0.653} & \underline{0.441} & \underline{124.0} & \textbf{25.85} & \textbf{0.707} & \textbf{0.373} & \textbf{73.7} \\
\midrule
\textbf{Average} & 22.48 & 0.657 & 0.461 & 559.2 & \underline{24.30} & \underline{0.722} & \underline{0.419} & \underline{389.2} & \textbf{25.56} & \textbf{0.741} & \textbf{0.390} & \textbf{229.8} \\
\bottomrule
\end{tabularx}
\end{table*}

\subsection{Mission Progress and GPU Memory}
\label{sec:execution-efficiency}

At Station Arch, AGE-Capture reaches both sides and the central region within $430.0\,\mathrm{s}$, whereas ActiveGS and TRACE remain near one side after $900\,\mathrm{s}$ (Fig.~\ref{fig:execution-efficiency}). Global scheduling directs the UAV across regions, while the reconstruction-frontier criterion avoids spending further capture time in regions with little remaining gain. Early replanning reduces waiting between successive trajectories by overlapping planning with flight. The broader trajectory coverage is consistent with the reconstruction quality improvements in Sec.~\ref{sec:reconstruction-quality}.

Across the ten scenes, AGE-Capture uses $0.69$--$3.26\,\mathrm{GiB}$ of peak GPU memory, averaging approximately one eleventh and one fifteenth of ActiveGS and TRACE, respectively (Table~\ref{tab:peak-gpu-memory}). The cumulative peak continues to rise for both baselines in the plotted runs, whereas it stabilizes shortly after initialization for AGE-Capture (Fig.~\ref{fig:execution-efficiency}). The renderability state allocates a bounded amount of storage per surface element. Repeated observations update that state in place; storage grows as new surface elements are added, rather than with the number of times a surface is observed.

\begin{table}[!t]
\centering
\small
\setlength{\tabcolsep}{3pt}
\caption{\MakeUppercase{Peak GPU Memory Usage} (\mbox{GiB}) \MakeUppercase{Across Online Active-Capture Systems. Lower Is Better.}}
\label{tab:peak-gpu-memory}
\begin{tabularx}{\linewidth}{>{\raggedright\arraybackslash}X>{\centering\arraybackslash}p{0.13\linewidth}>{\centering\arraybackslash}p{0.13\linewidth}>{\centering\arraybackslash}p{0.13\linewidth}}
\toprule
Scene & {\scriptsize ActiveGS} & TRACE & Ours \\
\midrule
Abandoned Warehouse & 15.85 & 22.39 & 1.01 \\
St Peter's Church & 19.77 & 32.09 & 1.58 \\
Barn & 13.63 & 14.25 & 0.96 \\
Garage Room & 16.56 & 28.48 & 1.04 \\
Station Arch & 14.83 & 20.04 & 1.15 \\
Graffiti Alley & 18.96 & 24.27 & 1.38 \\
Foug\`eres Castle & 11.19 & 17.65 & 0.69 \\
Rodez House & 15.66 & 20.83 & 2.00 \\
\'{E}vora Temple & 16.43 & 14.31 & 3.26 \\
Valve House & 16.43 & 21.77 & 1.33 \\
\midrule
\textbf{Average} & 15.93 & 21.61 & 1.44 \\
\bottomrule
\end{tabularx}
\end{table}

\begin{figure}[!htb]
  \centering
  \includegraphics[width=\columnwidth]{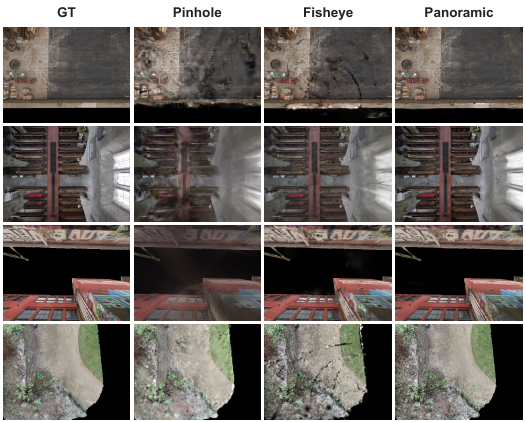}
  \caption{Qualitative reconstruction under different RGB camera models. Columns compare ground truth (GT), pinhole, fisheye, and panoramic observations; rows show Valve House, St Peter's Church, Graffiti Alley, and Foug\`eres Castle from top to bottom. Wider fields of view improve observation support for surfaces that are difficult to view under the fixed-pitch constraint.}
  \label{fig:rgb-camera-model-qualitative}
\end{figure}

\begin{figure*}[!t]
  \centering
  \begin{minipage}[c]{0.60\textwidth}
    \centering
    \includegraphics[width=\linewidth]{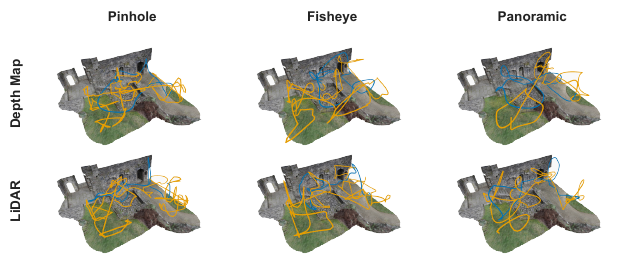}
    \centerline{(a)}
  \end{minipage}\hfill
  \begin{minipage}[c]{0.38\textwidth}
    \centering
    \includegraphics[width=\linewidth]{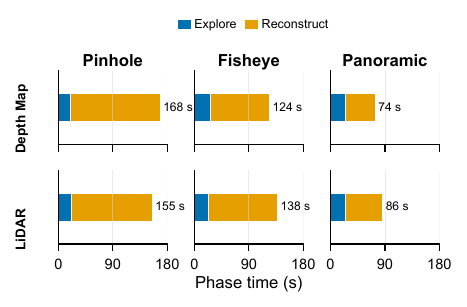}
    \centerline{(b)}
  \end{minipage}
  \caption{Exploration and reconstruction behavior under independently selected range and RGB configurations in Foug\`eres Castle. Rows compare depth-map and LiDAR geometry sensing; columns compare pinhole, fisheye, and panoramic RGB cameras. (a) UAV trajectories colored by exploration (blue) and reconstruction (orange). (b) Corresponding phase durations, with labels reporting total mission time.}
  \label{fig:sensor-phase-comparison}
\end{figure*}

\subsection{Sensor Configuration Adaptability}
\label{sec:sensor-configuration-adaptability}

We vary the RGB camera model and range-sensing modality independently, keeping the UniTriSplat reconstruction backend~\cite{zhu2026unitrisplat} and UAV motion constraints fixed. The RGB study compares pinhole, fisheye, and panoramic cameras in four scenes to evaluate how the field of view affects reconstruction-oriented capture. The range-sensor study combines depth-map and LiDAR inputs with all three RGB models in Foug\`eres Castle to examine whether dense and sparse geometric inputs affect RGB-side reconstruction capture. Table~\ref{tab:sensor-configurations} summarizes the corresponding sensor parameters.

Under identical attitude constraints, the panoramic configuration increases mean PSNR by $3.08\,\mathrm{dB}$ and reduces mean mission time from $559.2$ to $229.8\,\mathrm{s}$ relative to the pinhole configuration (Table~\ref{tab:rgb-camera-model-adaptability}). Figure~\ref{fig:rgb-camera-model-qualitative} shows that the improvements concentrate on ground and overhead structures that are difficult to observe under the fixed-pitch constraint. The wider field of view provides stronger single-view support for these surfaces, reducing the subsequent capture demand.

\begin{figure}[!b]
  \centering
  \includegraphics[width=\columnwidth]{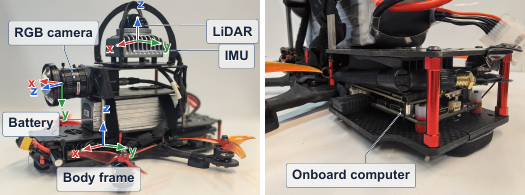}
  \caption{UAV hardware platform. Left: sensing payload, battery, and schematic coordinate axes. Right: onboard Jetson Orin NX computing unit.}
  \label{fig:uav-hardware-platform}
\end{figure}
\begin{table}[!t]
\centering
\small
\setlength{\tabcolsep}{1.5pt}
\caption{\MakeUppercase{Range-Input Adaptability in Foug\`eres Castle. Best and Second-Best Values Are Bold and Underlined, Respectively.}}
\label{tab:range-sensor-adaptability}
\begin{tabular}{>{\raggedright\arraybackslash}p{0.17\linewidth}>{\raggedright\arraybackslash}p{0.15\linewidth}>{\centering\arraybackslash}p{0.13\linewidth}>{\centering\arraybackslash}p{0.13\linewidth}>{\centering\arraybackslash}p{0.15\linewidth}>{\centering\arraybackslash}p{0.18\linewidth}}
\toprule
Range & RGB & PSNR$\uparrow$ & SSIM$\uparrow$ & LPIPS$\downarrow$ & Time (s)$\downarrow$ \\
\midrule
& Pinhole & 23.37 & 0.606 & 0.484 & 168.3 \\
Depth Map & Fisheye & 23.66 & 0.653 & 0.441 & 124.0 \\
& Panoramic & \underline{25.85} & \textbf{0.707} & \textbf{0.373} & \textbf{73.7} \\
\midrule
& Pinhole & 22.96 & 0.606 & 0.487 & 155.4 \\
LiDAR & Fisheye & 24.00 & 0.660 & 0.434 & 137.5 \\
& Panoramic & \textbf{26.05} & \underline{0.702} & \underline{0.377} & \underline{86.4} \\
\bottomrule
\end{tabular}
\end{table}

In Foug\`eres Castle, switching from depth maps to LiDAR changes PSNR by at most $0.41\,\mathrm{dB}$ when the RGB camera model is held fixed. The switch does not consistently shorten mission time (Table~\ref{tab:range-sensor-adaptability}). Figure~\ref{fig:sensor-phase-comparison} shows that different geometric inputs change exploration trajectories and the allocation of time between exploration and reconstruction. Although dense depth maps and sparse LiDAR affect geometric exploration differently, RGB-side capture continues to be organized by residual reconstruction value, yielding similar final quality with either input.

Panoramic RGB shortens the reconstruction phase but does not eliminate it under either range-sensing modality (Fig.~\ref{fig:sensor-phase-comparison}). This indicates that a wider field of view increases the initial observation support accumulated during exploration, while reconstruction subsequently supplements spatial and angular support for specific surfaces to meet reconstruction-frontier completion.

Across the tested configurations, the same capture loop uses geometric exploration to establish where the UAV can move and residual reconstruction gain to decide where further observations are needed.

\begin{figure*}[t]
  \centering
  \includegraphics[width=\textwidth]{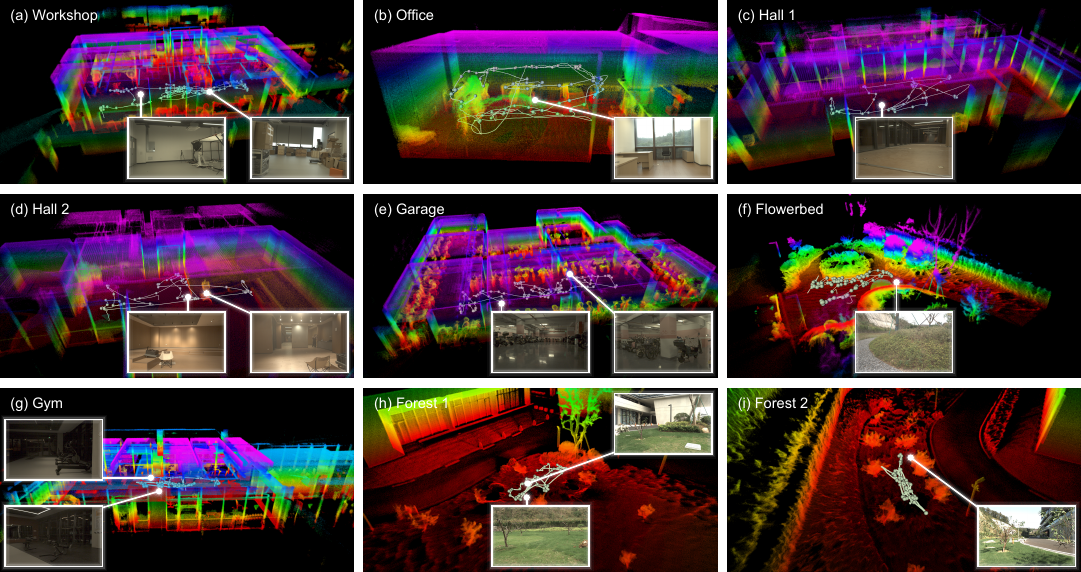}
  \caption{Nine real-world test environments. Each panel combines a LiDAR point-cloud map, the UAV trajectory, and event-frame acquisition positions. Insets show representative onboard RGB observations.}
  \label{fig:real-world-test-environments}
\end{figure*}
\begin{figure*}[t]
  \centering
  \includegraphics[width=\textwidth]{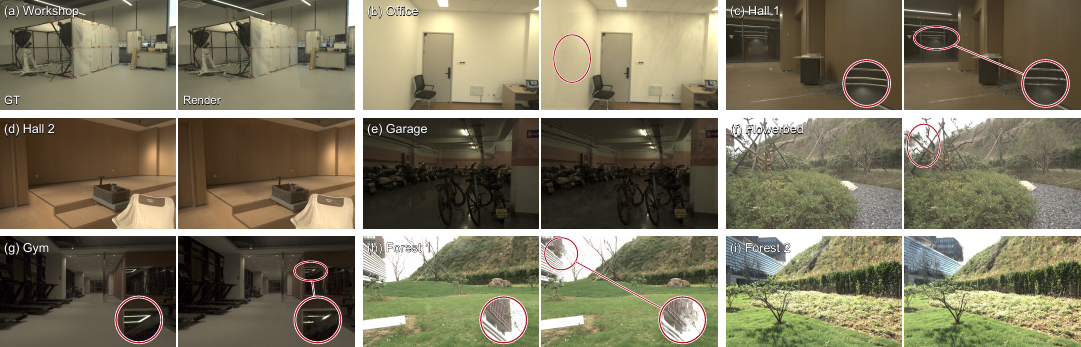}
  \caption{Real-world reconstruction examples. Each pair shows ground truth (GT, left) and our reconstruction (right). Red outlines identify regions with residual reconstruction artifacts. Circular insets show corresponding GT and reconstruction details at the same position and magnification.}
  \label{fig:real-world-reconstruction}
\end{figure*}
\subsection{Real-World Experiments}
\label{sec:real-world-experiments}

\subsubsection{Hardware Setup}

\begin{table}[t]
\centering
\small
\setlength{\tabcolsep}{4pt}
\renewcommand{\arraystretch}{1.08}
\caption{UAV platform and sensing configuration.}
\label{tab:real-platform}
\begin{tabularx}{\linewidth}{l>{\centering\arraybackslash}X}
\toprule
Component & Specification \\
\midrule
\textit{Platform} &  \\
Airframe & Carbon-fiber quadrotor \\
Onboard computer & NVIDIA Jetson Orin NX, 16 GB \\
Flight controller & CUAV Nora \\
Battery & 6S, 6000 mAh \\
\midrule
\textit{Sensing and acquisition} &  \\
LiDAR–IMU & Livox MID-360 \\
RGB camera & Hikrobot CA023-10UC \\
RGB resolution & 1920 × 1200 \\
RGB frame rate & 10 Hz \\
Fixed exposure & 2 ms \\
\bottomrule
\end{tabularx}
\end{table}

\begin{table}[t]
\centering
\small
\setlength{\tabcolsep}{4pt}
\renewcommand{\arraystretch}{1.08}
\caption{Motion constraints and active-capture settings.}
\label{tab:real-parameters}
\begin{tabularx}{\linewidth}{l>{\centering\arraybackslash}X}
\toprule
Setting & Value \\
\midrule
\textit{Motion and control} &  \\
Speed limit (m/s) & 1.0 \\
Acceleration limit (m/s²) & 0.5 \\
Yaw rate limit (rad/s) & 2.0 \\
Yaw acceleration limit (rad/s²) & 4.0 \\
Command publishing rate (Hz) & 50 \\
\midrule
\textit{Active-capture settings} &  \\
Candidate positions & 20 \\
Completion threshold & 0.2 \\
\bottomrule
\end{tabularx}
\end{table}

The quadrotor platform is shown in Fig.~\ref{fig:uav-hardware-platform}, and its hardware and sensing configuration are listed in Table~\ref{tab:real-platform}. All online modules run onboard the Jetson Orin NX, with FAST-LIO2 providing LiDAR--inertial odometry~\cite{xu2022fastlio2}. We calibrate camera intrinsics with the ROS camera\_calibration package,\footnote{\urlstyle{same}\url{https://index.ros.org/p/camera_calibration/}} and estimate LiDAR--camera extrinsics using the method of Koide et al.~\cite{koide2023calibration}. Sensor data are recorded in ROS bags for offline post-processing.

\subsubsection{Experimental Environments}
\label{sec:real-world-test-environments}

We conduct UAV active-capture experiments in nine real-world scenes, covering indoor environments such as a workshop, an office, halls, a garage, and a gym, as well as outdoor environments including a flowerbed and wooded areas (Fig.~\ref{fig:real-world-test-environments}). These scenes differ in spatial layout, obstacle distribution, and occlusion conditions.

\subsubsection{Post-processing Workflow}

We sample the RGB sequence at $2\,\mathrm{Hz}$ while retaining all event frames, and initialize GLUEMAP with camera poses from FAST-LIVO2 so that pose optimization starts at the original LiDAR-map scale~\cite{fastlivo2,gluemap}. We then estimate a global similarity transform from corresponding camera centers to align the LiDAR map with the optimized visual coordinate system~\cite{umeyama1991}.

Hidden point removal (HPR) identifies visible LiDAR points, which are projected to generate $480\times300$ depth maps~\cite{hpr2007}. Visible LiDAR points augment the sparse visual point cloud for Gaussian initialization, while depth supervision constrains reconstruction of weakly textured surfaces. The 3DGS model~\cite{kerbl20233dgaussians} is optimized for 30,000 iterations using event-frame RGB images and valid LiDAR depths.

For each scene, we randomly select a fixed set of 100 non-event images and report mean PSNR, SSIM, and LPIPS at the original resolution. In Forest 2, the candidate pool excludes positions within $0.2\,\mathrm{m}$ of the initial camera and retains one image per $0.2\,\mathrm{m}$ spatial voxel before sampling. These images contribute to upstream camera-pose estimation and visual initialization but do not participate in Gaussian optimization of the event-frame model.

\subsubsection{Reconstruction Performance}
\label{sec:real-world-reconstruction-performance}

We acquired 49--115 event frames per scene, with $132.2$--$278.9\,\mathrm{s}$ between the first and last events (Table~\ref{tab:real-world-reconstruction}). The GT--render comparisons in Fig.~\ref{fig:real-world-reconstruction} illustrate the resulting reconstructions: indoor scenes retain the main spatial layout, equipment contours, and surface appearance, while outdoor scenes recover the overall form of vegetation, ground surfaces, and surrounding buildings.

Outdoor scenes such as Flowerbed and Forest 1 contain rich vegetation textures and irregular edges that are harder to recover fully from limited observations. Local blur and contour misalignment in these high-frequency details increase image errors, contributing to their relatively low PSNR and SSIM. The branch and background details in the figure illustrate these errors.

Some distant surfaces in Flowerbed, Forest 1, and Hall 1 lie outside the near-field capture target and receive weaker observation support, leaving contour smearing or ghosting. Reflections and view-dependent appearance require more extensive observation support; under the relatively relaxed completion setting used to accommodate limited flight time, local artifacts remain, including wall mottling in Office and deformation of reflected light strips in Gym.

\subsubsection{Capture Stability}

\begin{figure}[!tb]
  \centering
  \includegraphics[width=\columnwidth]{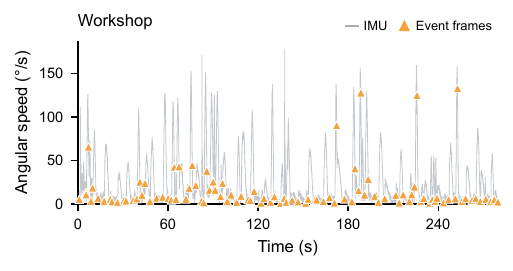}
  \centerline{(a)}
  \vspace{1mm}
  \includegraphics[width=\columnwidth]{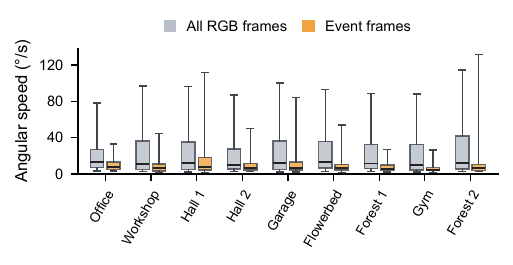}
  \centerline{(b)}
  \caption{Angular speed at event-frame acquisition during real-world missions. (a) Measured IMU angular speed and event-frame acquisition times over the Workshop mission. (b) Angular-speed distributions of all RGB frames and event frames within the same reconstruction acquisition interval across nine scenes. Frame-wise angular speed is the norm of the three-axis IMU angular velocity interpolated at each image timestamp. Boxes show the interquartile range, horizontal lines denote medians, and whiskers span the 5th--95th percentile range.}
  \label{fig:capture-angular-velocity}
\end{figure}

Our asynchronous capture mechanism estimates the target observation time from the time-parameterized trajectory and triggers event-frame updates through timestamps. As shown in Fig.~\ref{fig:capture-angular-velocity}, most event frames in Workshop are acquired at low angular speeds; across all nine scenes, their median angular speed is $32.1\%$--$53.3\%$ lower than that of all RGB frames acquired during the same interval. With the fixed $2\,\mathrm{ms}$ exposure, lower acquisition angular speed helps reduce rotational displacement during exposure and the associated risk of motion blur.

\begin{table}[t]
\centering
\small
\setlength{\tabcolsep}{3pt}
\renewcommand{\arraystretch}{1.08}
\caption{Real-world reconstruction with event-frame counts and capture durations (first-to-last event). Metrics are averaged over 100 non-event images per scene.}
\label{tab:real-world-reconstruction}
\begin{tabularx}{\linewidth}{l>{\centering\arraybackslash}X >{\centering\arraybackslash}X >{\centering\arraybackslash}X >{\centering\arraybackslash}X >{\centering\arraybackslash}X}
\toprule
Scene & Time (s) & Frames & PSNR\ensuremath{\uparrow} & SSIM\ensuremath{\uparrow} & LPIPS\ensuremath{\downarrow} \\
\midrule
Workshop & 278.9 & 98 & 27.16 & 0.875 & 0.317 \\
Office & 183.8 & 66 & 24.37 & 0.907 & 0.308 \\
Hall 1 & 132.2 & 58 & 28.74 & 0.777 & 0.465 \\
Hall 2 & 138.9 & 49 & 25.39 & 0.771 & 0.440 \\
Garage & 256.4 & 115 & 28.96 & 0.768 & 0.529 \\
Flowerbed & 205.9 & 91 & 21.40 & 0.640 & 0.297 \\
Gym & 160.0 & 61 & 24.43 & 0.541 & 0.559 \\
Forest 1 & 142.9 & 54 & 21.79 & 0.660 & 0.290 \\
Forest 2 & 144.1 & 62 & 18.30 & 0.557 & 0.297 \\
\bottomrule
\end{tabularx}
\end{table}

A small number of event frames are still acquired at high angular speeds; in Hall 1 and Forest 2, the event-frame 95th percentile exceeds that of all RGB frames despite the lower median. These cases may be associated with computational blocking or timing-estimation errors during long-trajectory execution, causing the predicted acquisition time to deviate from the actual low-angular-speed phase.

\subsubsection{Onboard Performance}

To accommodate the UAV's limited flight time, we set the candidate count to $K=20$ and the completion threshold to $\tau_{\mathrm{fine}}=0.2$ (Table~\ref{tab:real-parameters}). The remaining active-capture parameters are unchanged from simulation. In Forest 2, renderability-state updates and complete NBV planning take median times of $85.5\,\mathrm{ms}$ and $925.3\,\mathrm{ms}$, respectively, with corresponding 95th-percentile times of $104.7\,\mathrm{ms}$ and $1022.9\,\mathrm{ms}$. Although these computations take longer than in simulation, observation updates run independently of planning, and the UAV continues flying while the next plan is computed.
\begin{figure*}[!t]
  \centering
  \includegraphics[width=\textwidth]{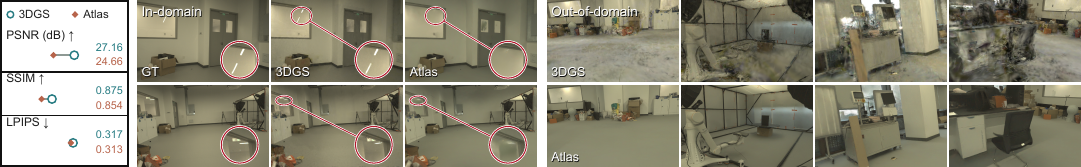}
  \caption{Workshop reconstruction from AGE-Capture event frames. Both outputs use a 3D Gaussian representation. ``3DGS'' denotes scene-specific optimization, whereas ``Atlas'' denotes world-model-based reconstruction. The metrics are averaged over 100 held-out views. The insets compare reflected details against the reference images, and the right panels show four extrapolated views.}
  \label{fig:world-model-comparison}
\end{figure*}
\subsection{Ablation Studies}
\label{sec:ablation-studies}

We ablate continuous capture and sweep the reconstruction-frontier threshold in Barn, St Peter’s Church, and Foug\`eres Castle. Both continuous-capture variants and threshold sweeps run without the $900\,\mathrm{s}$ limit used in the main benchmark. Continuous-capture variants use the same reconstruction-driven completion condition.

\subsubsection{Continuous Capture}

\begin{figure}[!tb]
  \centering
  \includegraphics[width=\columnwidth]{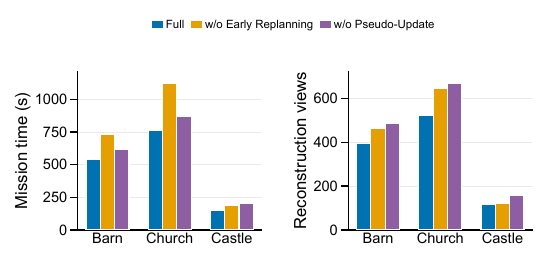}
  \vspace{-1.5mm}
  \begin{minipage}{0.49\columnwidth}\centering (a)\end{minipage}\hfill%
  \begin{minipage}{0.49\columnwidth}\centering (b)\end{minipage}
  \caption{Ablation of continuous capture for the full system, without early replanning, and without pseudo-update. (a) Mission time required to reach the same reconstruction-driven completion condition. (b) Corresponding number of reconstruction views. The $900\,\mathrm{s}$ limit is not applied.}
  \label{fig:continuous-capture-ablation}
\end{figure}
\begin{figure}[!tb]
  \centering
  \includegraphics[width=\columnwidth,trim=0 7bp 0 0,clip]{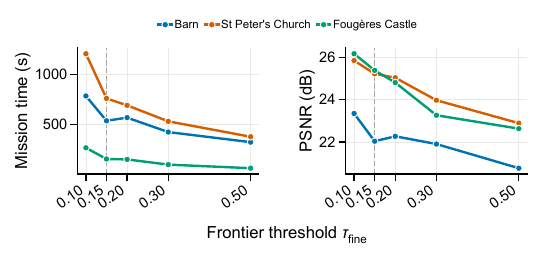}
  \vspace{-4mm}
  \caption{Sensitivity to the reconstruction-frontier completion threshold $\tau_{\mathrm{fine}}$. The left and right plots report mission time and offline 3DGS reconstruction PSNR, respectively. Smaller thresholds impose stricter completion conditions; $0.15$ is the simulation default.}
  \label{fig:completion-threshold-sensitivity}
\end{figure}

Continuous-capture ablations separate two sources of inefficiency in closed-loop capture: waiting caused by serial planning and motion, and repeated observations caused by stale state. Under the same reconstruction-driven completion criterion, disabling early replanning increases total mission time across the three scenes by $40.7\%$ (Fig.~\ref{fig:continuous-capture-ablation}(a)). Early replanning completes the subsequent decision during execution of the active trajectory, overlapping planning and motion.

The pseudo-update ablation reveals the effect of stale state on capture decisions. Across the three scenes, disabling pseudo-update increases the total number of matched RGB--pose observations written to the reconstruction backend by $27.3\%$ and total mission time by $16.7\%$ (Fig.~\ref{fig:continuous-capture-ablation}). Before the preceding target produces a real observation, its expected surface support is absent from the persistent state, so subsequent planning can select overlapping observations. Pseudo-update incorporates this expected support into prospective scoring, reducing repeated capture between successive decisions.

\subsubsection{Reconstruction-Driven Completion}

The reconstruction-frontier threshold $\tau_{\mathrm{fine}}$ controls the residual reconstruction value accepted at task completion. We evaluate $\tau_{\mathrm{fine}}\in\{0.10, 0.15, 0.20, 0.30, 0.50\}$; smaller values require the reconstruction-frontier value of every remaining region to fall to a lower level.

Across all three scenes, increasing $\tau_{\mathrm{fine}}$ from $0.10$ to $0.50$ shortens missions but lowers offline reconstruction PSNR (Fig.~\ref{fig:completion-threshold-sensitivity}). Earlier completion leaves less observation support for offline reconstruction.

Relative to $\tau_{\mathrm{fine}}=0.10$, the simulation default of $0.15$ reduces mission time by $31.7\%$--$42.4\%$ across the three scenes, with a PSNR decrease of $0.63$--$1.31\,\mathrm{dB}$.

\section{Discussion}
\label{sec:discussion}

To fit each real-world mission within a single flight, we limit the capture region and relax the completion criterion, leaving some distant surfaces with insufficient observations. In the threshold ablation, stricter completion criteria require more capture time (Fig.~\ref{fig:completion-threshold-sensitivity}). R3CON shows that renderability-driven capture can achieve high-quality reconstruction with greater freedom in viewpoint selection~\cite{r3con}. Reconstructing details such as reflections also depends on the backend's ability to represent view-dependent appearance.

In the Workshop scene, we use the same 98 event frames for scene-specific 3DGS optimization and reconstruction with Atlas\footnote{\urlstyle{same}\url{https://atlas-beta.worldlabs.ai/}}. The exported Atlas model uses only degree-zero spherical harmonics, limiting its ability to reproduce view-dependent reflections. Missing reflected details and overall color shifts relative to the reference images (Fig.~\ref{fig:world-model-comparison}) are consistent with Atlas scoring lower than scene-specific 3DGS on both PSNR and SSIM. Atlas nevertheless produces more coherent textures and object contours in the illustrated regions, and the scene remains recognizable in all four extrapolated views. The different reconstructions obtained from the same observations motivate calibrating the completion threshold to the reconstruction backend and the target quality.

For a UAV with limited attitude control, a wider field of view can reduce the need to reposition by making more viewing directions available from a single pose (Table~\ref{tab:rgb-camera-model-adaptability}). Matching the sensing configuration to the platform's attitude capabilities can therefore improve observation support within a fixed flight duration.

Compared with ground mobile platforms, UAV deployment places particular demands on payload, onboard computation, and endurance, while flight-safety requirements further restrict the accessible capture space. The real-world experiments demonstrate that a reconstruction-driven autonomous capture loop can operate under these constraints.

\section{Conclusion}
\label{sec:conclusion}

This paper presented AGE-Capture, extending reconstruction-aware viewpoint decisions into an autonomous capture loop for real UAVs.
Simulation and onboard experiments show that a lightweight renderability state can provide a common basis for sensor adaptation, observation selection, and task-completion decisions, allowing acquisition to advance autonomously according to residual reconstruction value; coupled with continuous execution, this closed loop operates under limited onboard computation and flight resources.
Future work will adapt the framework to different robotic platforms and extend it to cooperative multi-robot and multi-session capture, supporting sustained autonomous reconstruction over larger areas.

\IEEEtriggeratref{38}
\bibliographystyle{IEEEtran}
\bibliography{references}
\end{document}